\documentclass[journal]{IEEEtran}
\usepackage{amsmath,amsfonts}
\usepackage{algorithmic}
\usepackage{algorithm}
\usepackage{array}
\usepackage[caption=false,font=normalsize,labelfont=sf,textfont=sf]{subfig}
\usepackage{textcomp}
\usepackage{stfloats}
\usepackage{url}
\usepackage{verbatim}
\usepackage{graphicx}
\usepackage{cite}
\usepackage{hyperref}
\usepackage{url}

\usepackage{graphicx}
\usepackage{subcaption}
\usepackage{multirow}
\usepackage[table,xcdraw]{xcolor}
\usepackage[most]{tcolorbox}
\usepackage{booktabs}
\usepackage{bbding}
\usepackage{wrapfig}
\usepackage{algorithm}
\usepackage{algorithmic}
\usepackage{enumitem}

\definecolor{mypink}{RGB}{216, 18, 126}
\definecolor{citeblue}{RGB}{1, 113, 188}

\hypersetup{
    colorlinks=true,
    linkcolor=mypink,
    urlcolor=mypink,
    citecolor=citeblue,
}

\newif\ifshowrev
\showrevfalse % 隐藏修订

\definecolor{revblue}{RGB}{0, 92, 175}

\ifshowrev
  \newcommand{\rev}[1]{\textcolor{revblue}{#1}}
\else
  \newcommand{\rev}[1]{#1}
\fi

\begin{document}

\title{Adapting LLMs to Time Series Forecasting via \\ Temporal Heterogeneity Modeling and Representation Alignment}

%\iffalse
\author{
    Yanru Sun, 
    Emadeldeen Eldele, 
    Zongxia Xie$^\dag$, 
    Yucheng Wang, \\
    Wenzhe Niu, 
    Qinghua Hu, 
    Chee Keong Kwoh, 
    Min Wu$^\dag$

%\thanks{Received 11 March 2025; revised 11 August 2025; accepted 1 December 2025. Date of publication 11 December 2025; date of current version 3 June 2026. 
%This work was supported in part by the National Natural Science Foundation of China under Grants 625B2126, U23B2049, 62376194, and 61925602; in part by the China Scholarship Council under Grant 202406250137; and in part by AIMfg, A$^*$STAR, Singapore, under its MTI WS Fund (W24MCMF012).} %Singapore Ministry of Trade and Industry under Grant W24MCMF012.}
\thanks{Yanru Sun, Zongxia Xie, Wenzhe Niu, and Qinghua Hu are with Tianjin University. (e-mail: yanrusun@tju.edu.cn, caddiexie@hotmail.com, niuwenzhe@tju.edu.cn, and huqinghua@tju.edu.cn)}
\thanks{Emadeldeen Eldele is with the Department of Computer Science, Khalifa University, Abu Dhabi, UAE. (e-mail: emad0002@ntu.edu.sg)}
\thanks{Yucheng Wang is with Institute for Infocomm Research, A*STAR, Singapore and the School of Electrical and Electronic Engineering, Nanyang Technological University, Singapore (Email: yucheng003@e.ntu.edu.sg)}
\thanks{Chee Keong Kwoh is with Nanyang Technological University. (e-mail: ASCKKWOH@ntu.edu.sg)}
\thanks{Min Wu is with the Institute for Infocomm Research, A*STAR. (e-mail: wumin@a-star.edu.sg)}
\thanks{$^\dag$Corresponding authors: Zongxia Xie and Min Wu.}}
%\fi

\iffalse
\author{IEEE Publication Technology,~\IEEEmembership{Staff,~IEEE,}
        % <-this % stops a space
\thanks{This paper was produced by the IEEE Publication Technology Group. They are in Piscataway, NJ.}% <-this % stops a space
\thanks{Manuscript received April 19, 2021; revised August 16, 2021.}}
\fi

\markboth{Journal of \LaTeX\ Class Files,~Vol.~14, No.~8, August~2021}%
{Shell \MakeLowercase{\textit{et al.}}: A Sample Article Using IEEEtran.cls for IEEE Journals}


\maketitle

\begin{abstract}
%Large Language Models (LLMs) have recently demonstrated impressive performance in natural language processing due to their strong generalization and sequence modeling capabilities.
%However, their direct application to time series forecasting remains challenging due to two fundamental issues: the inherent heterogeneity of temporal patterns and the modality gap between continuous numerical signals and discrete language representations.
Recent advances have demonstrated that Large Language Models (LLMs) can be effectively adapted for time series forecasting, revealing strong potential beyond natural language tasks.
However, their performance remains constrained by two fundamental challenges: the inherent heterogeneity of temporal patterns and the modality gap between continuous numerical signals and discrete language representations.
In this work, we propose \textbf{TALON} (Temporal-heterogeneity And Language-Oriented Network), a unified framework that enhances LLM-based forecasting by modeling temporal heterogeneity and promoting representation alignment.
Specifically, we design a Heterogeneous Temporal Encoder that partitions multivariate time series into structurally coherent segments, enabling localized expert modeling across diverse temporal patterns.
To bridge the modality gap, we introduce a Representation Alignment Module that projects temporal features toward LLM-compatible representations, making time series more amenable to LLMs and thereby unlocking their modeling potential, while eliminating the need for handcrafted prompts during inference.
Extensive experiments on seven real-world benchmarks demonstrate that TALON achieves superior performance across all datasets, with average MSE improvements of up to 11\% over recent state-of-the-art methods, while maintaining higher efficiency.
These results underscore the effectiveness of incorporating both pattern-aware modeling and representation-level alignment when adapting LLMs for time series forecasting.
%The code is available at: \url{https://anonymous.4open.science/r/TALON-2B48}.
The code is available at: \url{https://github.com/syrGitHub/TALON}.
\end{abstract}


\section{Introduction}
\IEEEPARstart{T}{ime} series forecasting plays a critical role in a wide range of real-world applications, spanning high-stakes domains such as healthcare monitoring \cite{jin2023uncertainty, yan2023towards} and power grid control \cite{shao2024exploring}, as well as everyday services including weather forecasting \cite{sun2021solar, sun2022accurate}, traffic prediction \cite{jin2024spatial}, and energy load estimation \cite{wu2024enhanced}.
To ensure reliable forecasting in such complex and dynamic environments, it is essential to effectively model long-range temporal dependencies \cite{sun2025hierarchical, liu2024itransformer, yang2025temporal, gong2026symbolic}.

%Recently, large language models (LLMs) have demonstrated remarkable generalization and representation capabilities across a wide range of language and vision tasks \cite{touvron2023llama, achiam2023gpt, team2024qwen2, liu2024sora}.
%Inspired by the shared sequential nature of time series and language data, recent research has explored LLMs as general-purpose forecasters for time series applications \cite{ansari2024chronos, jin2024timellm, liu2024autotimes}, aiming to leverage their strong sequence modeling capabilities.

Recently, large language models (LLMs) have demonstrated remarkable generalization and sequence modeling capabilities across diverse tasks \cite{achiam2023gpt, team2024qwen2, zhao2026luve, yu2026visual}.
Beyond linguistic semantics, the effectiveness of LLMs is largely attributed to long-context modeling, stable autoregressive generation, and robust sequence learning through large-scale pretraining.
Such properties are also essential to time series forecasting, particularly for capturing long-term dependencies and ensuring stable multi-step prediction, making LLMs well suited for this domain \cite{ansari2024chronos, jin2024timellm, liu2024autotimes}.

\begin{figure}[t]
    \centering
    \includegraphics[width=1.0\linewidth]{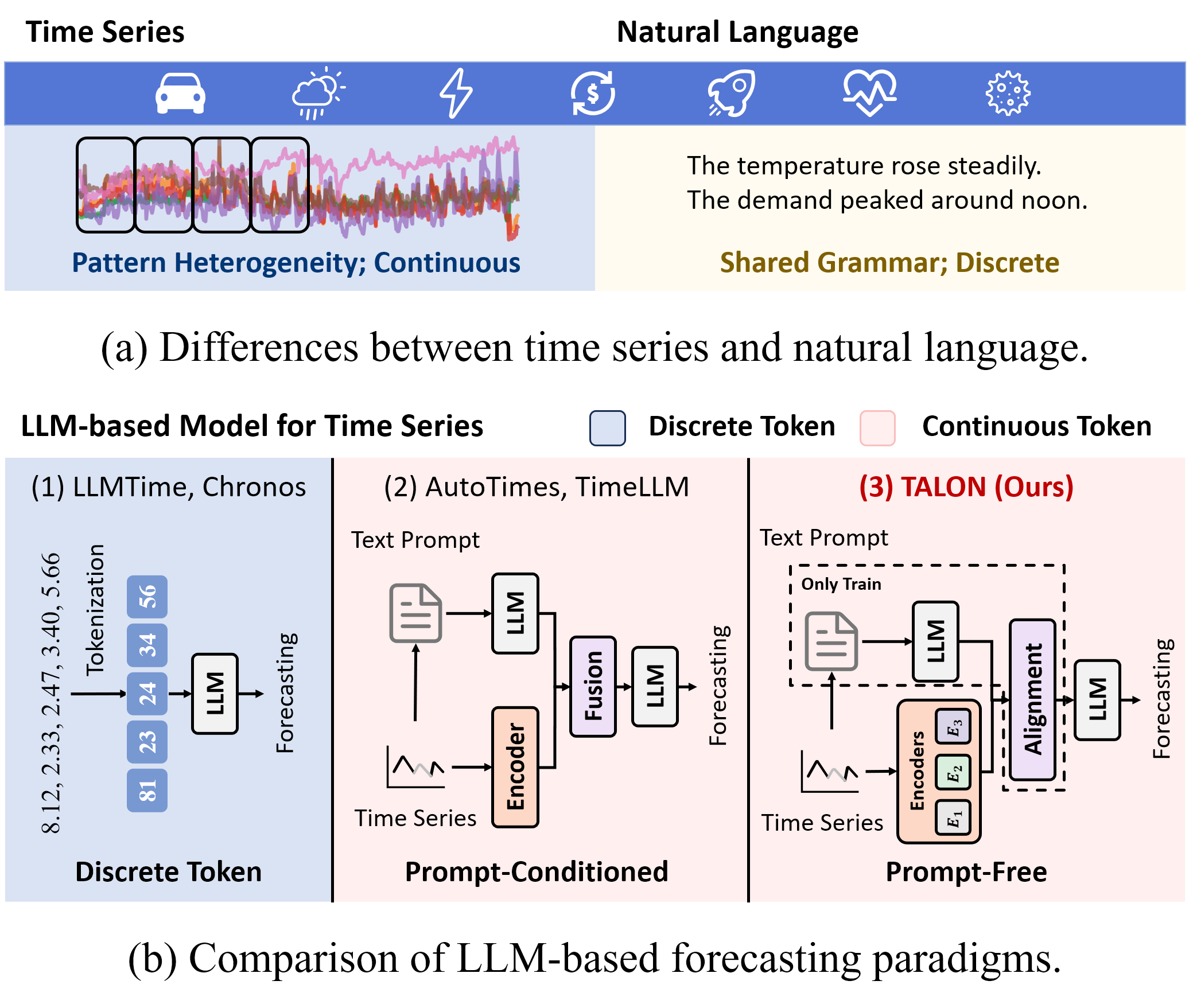}
    \caption{(a) Time series are continuous and structurally diverse, whereas natural language is discrete and syntactically uniform, posing a modality gap that hinders the direct application of LLMs to time series forecasting. (b) Our proposed TALON introduces a framework that integrates heterogeneous temporal encoding with contrastive representation alignment, enabling pattern-aware and representation-grounded forecasting without relying on prompts during inference.}
    \label{fig:problem}
\end{figure}

%However, as illustrated in Figure~\ref{fig:problem}~(a), multivariate time series often exhibit intrinsic heterogeneity, where different segments and variables follow diverse and evolving patterns \cite{qiu2025duet, liu2025moiraimoe}.
%In contrast, LLMs are pretrained on text corpora with globally consistent grammatical structures, which limits their ability to handle fragmented or nonstationary temporal inputs.
%Moreover, time series are continuous and real-valued, governed by strong temporal dependencies, whereas LLMs are inherently designed for discrete, symbolic sequences \cite{ansari2024chronos}.
%This discrepancy in both structure and modality poses significant challenges for directly applying LLMs to time series forecasting \cite{liu2025timecma}.

However, as illustrated in Figure~\ref{fig:problem}~(a), time series differ fundamentally from language in that they are continuous and real-valued, rather than discrete and symbolic \cite{ansari2024chronos}. 
In addition, multivariate time series often exhibit intrinsic heterogeneity, where different segments and variables follow diverse and evolving patterns, leading to nonstationary and locally varying dynamics \cite{qiu2025duet, liu2025moiraimoe}. 
This discrepancy in both modality and temporal structure poses significant challenges for directly applying LLMs to time series forecasting \cite{liu2025timecma}.

Motivated by these challenges, existing LLM-based forecasting methods generally aim to transform time series representations into forms that are more compatible with the representation space expected by LLMs.
As illustrated in Figure~\ref{fig:problem}~(b), these approaches can be broadly categorized into two paradigms:
(1) Tokenization-based methods, which discretize continuous sequences into symbolic tokens \cite{gruver2023large, ansari2024chronos}; and
(2) Prompt-conditioned methods, which prepend handcrafted textual templates to time series inputs \cite{liu2024autotimes, jin2024timellm}.
While both paradigms attempt to adapt LLMs to time series data, they fail to fully account for the modality gap. Specifically, they either disrupt temporal continuity, discard fine-grained numerical structure, or suffer from weak alignment and a reliance on handcrafted prompts.

%To address these challenges, we propose \textbf{TALON} (Temporal-heterogeneity And Language-Oriented Network), a unified framework that bridges the modality gap by jointly modeling temporal heterogeneity and enforcing semantic alignment between time-series and language representations.
To address these challenges, we propose \textbf{TALON} (Temporal-heterogeneity And Language-Oriented Network), a unified framework that bridges the modality gap from an LLM-centric perspective by leveraging LLMs as powerful sequence learners.
First, we propose a \textbf{Heterogeneous Temporal Encoder (HTE)} to partition multivariate time series into structurally homogeneous segments based on their statistical and temporal properties, enabling pattern-aware expert modeling.
Second, we introduce a \textbf{Representation Alignment Module (RAM)} that aligns continuous temporal features with LLM-compatible embeddings in a shared representation space, eliminating the need for handcrafted prompts and facilitating effective integration with pretrained LLMs.
Finally, we employ a \textbf{LLM Forecasting Head (LFH)} that combines a pretrained LLM with lightweight projection layers to autoregressively generate future segments from the aligned representations.
We evaluate TALON on seven real-world time series forecasting benchmarks, where it consistently outperforms both LLM-based and deep learning baselines across various prediction horizons.
Our contributions are summarized as follows:

\begin{itemize}
    \item We identify and characterize the modality misalignment problem in LLM-based time series forecasting from both structural and representation perspectives, highlighting how the discrepancy between continuous signals and discrete language inputs limits existing paradigms.
    \item We propose TALON, a novel framework that integrates heterogeneous pattern decomposition and representation alignment to enable fine-grained forecasting and cross-modal representation learning.
    \item Experimentally, TALON consistently outperforms state-of-the-art baselines across seven real-world forecasting benchmarks, achieving up to 11\% reduction in MSE while improving both accuracy and generalization.
\end{itemize}

\section{Related Work}
\textbf{Deep Learning for Time Series Forecasting.}
% deep learning method patchtst, itransformer, timesnet, pattern: linear-moe, duet, tfps, mofe-time
Deep learning has become a cornerstone in time series forecasting, with various architectures designed to capture complex temporal dependencies \cite{xu2025toward, xu2025twin}. 
Convolutional neural networks are widely used to extract local temporal patterns and variable-wise dependencies \cite{wu2023timesnet, eldele2024tslanet, wang2025timemixer}. 
More recently, Transformer-based models have gained popularity due to their global receptive fields and self-attention mechanisms, which enable long-range dependency modeling. 
For instance, PatchTST \cite{nie2023time} proposes a channel-independent patching mechanism to decouple variable interactions, while iTransformer \cite{liu2024itransformer} enhances multivariate modeling by treating each univariate series as an individual token.
Building on these ideas, recent network architectures further capture complex temporal patterns, including multiscale recurrent networks with scale attention \cite{guo2023multivariate} and residual multiscale TCN sparse expert networks combined with Informer \cite{bao2025long}. These approaches effectively model both short- and long-term dependencies while addressing heterogeneity and non-stationarity in multivariate time series.
To further handle heterogeneous temporal patterns, several methods introduce mechanisms such as mixture-of-experts \cite{ni2024mixture, liu2025freqmoe} and subspace-based pattern grouping \cite{sun2025learning}, improving robustness to diverse dynamics. 
Despite these advances, most existing methods remain constrained by limited parameterization and small-scale training corpora \cite{cai2024modeling, liu2024lighttr}.

\textbf{Large Language Models for Time Series.}
Motivated by the sequential nature shared between time series and language, such as local-to-global dependency structures and autoregressive generation, recent studies have explored adapting LLMs to time series forecasting \cite{gruver2023large, jin2024position}.
One line of work discretizes time series into symbolic tokens via quantization or pattern clustering, enabling direct utilization of token-based LLMs \cite{gruver2023large, ansari2024chronos}.
Another line of research retains raw numerical inputs and leverages textual prompts to provide contextual guidance \cite{liu2024autotimes, jin2024timellm, niu2025langtime}. 
%While these approaches benefit from the generalization capabilities of pretrained LLMs, they typically overlook the mismatch in both temporal structure and representation space between natural language and continuous time series, leading to limited scalability and suboptimal representation alignment.
Beyond forecasting-oriented adaptation, more recent studies have begun to explore broader time-series reasoning and multimodal integration with large models. For example, TimeOmni-1 \cite{guan2026timeomni} extends LLMs toward unified time-series reasoning through multi-stage training and reasoning-oriented task design, while OpenTSLM \cite{langer2025opentslm} incorporates time series as a native modality for text-time-series reasoning in medical settings.
Despite their differences in task scope and modeling objective, these lines of research still face the broader challenge of bridging continuous temporal signals with language-oriented representation spaces, which can limit scalability and hinder effective cross-modal alignment.
%These works highlight a broader shift from task-specific forecasting adaptation toward more general temporal understanding and reasoning. In contrast, our work focuses on long-term forecasting, where the key challenge is to efficiently adapt continuous and heterogeneous temporal signals to pretrained LLMs through pattern-aware temporal encoding and lightweight representation alignment.

\begin{figure*}
    \centering
    \includegraphics[width=1.0\linewidth]{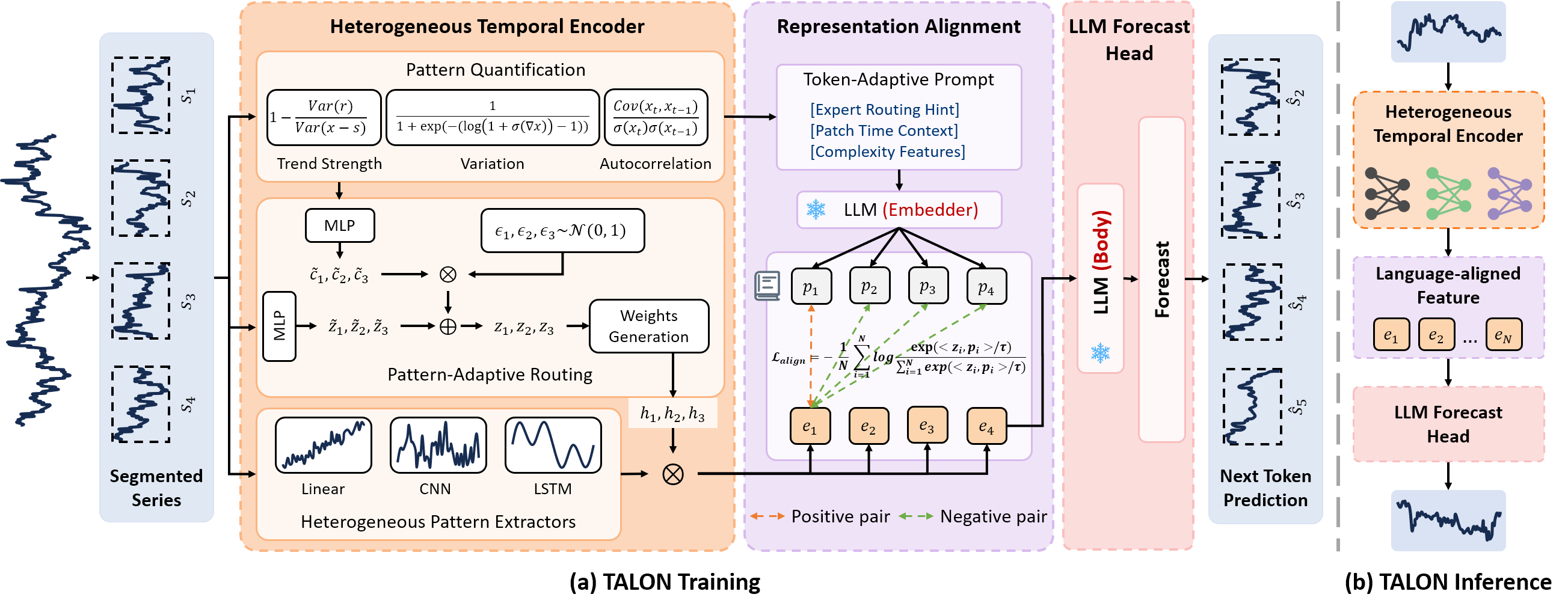}
    \caption{Overview of the TALON architecture. (1) Heterogeneous Temporal Encoder estimates coarse segment-level complexity cues and routes each segment to specialized experts via a pattern-based routing mechanism. (2) Representation Alignment Module generates structured, token-level prompts that encode expert routing hints and temporal context, and applies contrastive learning to align time-series features with LLM-compatible representations, thereby enabling stable and effective integration without inference-time prompts. (3) LLM Forecasting Head takes the aligned features as input and performs autoregressive next-segment prediction. This design supports pattern-aware modeling, prompt-free inference, and representation-aligned forecasting under heterogeneous temporal patterns.}
    \label{fig:model}
\end{figure*}

\section{Preliminaries}
Given a multivariate input sequence $X = (x_{t-L+1}, \dots , x_t) \in \mathbb{R}^{L \times C}$, the goal of time series forecasting is to predict the future values $Y = (x_{t+1}, \dots , x_{t+H}) \in \mathbb{R}^{H \times C}$, where $L$ is the look-back window length, $H$ is the forecasting horizon, and $C$ is the number of variables. The task is to learn a predictive function $f_{\theta}$ such that $Y = f_{\theta}(X)$.  

\section{Method}
\label{sec:Method}
\subsection{Overall Architecture}
As illustrated in Figure~\ref{fig:model}, our proposed framework \textbf{TALON} consists of three key components: the Heterogeneous Temporal Encoder (HTE), the Representation Alignment Module (RAM), and the LLM Forecasting Head (LFH).

To focus on modeling temporal variations, we follow the channel-independent strategy \cite{liu2024autotimes}, decomposing the multivariate input into $C$ separate univariate sequences.
Each univariate sequence is further segmented into $N$ consecutive non-overlapping patches of length $S$, with each patch denoted as $s_i = \{x_{(i-1)S+1}, \dots, x_{iS}\} \in \mathbb{R}^{S}, i=1, \cdots, N$.

The HTE module extracts lightweight token-level statistical cues from each patch and dynamically routes it to a specialized expert (e.g., Linear, CNN, LSTM) via a learnable gating mechanism, enabling localized and pattern-aware temporal modeling.
Next, the RAM constructs token-adaptive prompts based on complexity-related routing cues and temporal context. These prompts are processed by a frozen LLM to produce LLM-compatible representations in the language embedding space.
To bridge the modality gap between continuous time-series features and discrete language representations, we introduce a fine-grained contrastive alignment loss at the token level \cite{jin2025time}. This encourages the time-series-derived representations to align closely with the language embeddings, effectively transforming them into language-aligned features suitable for LLM-based forecasting.
Finally, the LFH takes the aligned embeddings as input and employs an autoregressive decoder, consisting of a frozen LLM and a linear projection layer, to generate forecasting outputs. This design supports variable prediction lengths while maintaining low inference cost.
We elaborate on each module in the following subsections.

\subsection{Heterogeneous Temporal Encoder}
Multivariate time series often exhibit complex and heterogeneous temporal dynamics, including diverse trends, fluctuations, and long-range dependencies across variables and time \cite{shao2024exploring}. To effectively model such variability, we propose the Heterogeneous Temporal Encoder (HTE), which learns pattern-aware representations by dynamically adapting expert selection to coarse complexity-related cues and temporal structure of each input patch.

As shown in Figure~\ref{fig:model}, HTE consists of three key components: (1) Pattern Quantification, (2) Pattern-Adaptive Routing, and (3) Heterogeneous Pattern Extractors. 

\textbf{Pattern Quantification.}
%To characterize the local temporal structure of each patch, HTE computes a compact set of interpretable token-level statistical features (e.g., trend strength, variation, and autocorrelation). These features quantify local temporal dynamics and serve as the basis for routing each patch to a specialized modeling branch tailored to distinct temporal behaviors.
To provide lightweight and interpretable cues for routing, HTE represents each patch using three token-level statistical descriptors: trend strength, variation, and autocorrelation \cite{qiu2024tfb, li2024foundts}. These complementary descriptors capture low-frequency directional structure, short-term fluctuation intensity, and local temporal dependency, respectively. Although richer statistics are possible, we prioritize simplicity and robustness, and empirically find this compact descriptor set sufficient for effective routing.

For a univariate patch $s_i \in \mathbb{R}^S$, the resulting quantification vector is defined as $c_i = [c_1, c_2, c_3] \in \mathbb{R}^3$, where $c_1$, $c_2$, and $c_3$ denote trend strength, local variation, and autocorrelation coefficient, respectively. This vector provides a coarse description of the local temporal behavior of $s_i$ and serves as the input to the expert routing mechanism. The specific calculation formulas are provided in the Appendix~\textcolor{mypink}{D}.%\ref{sec_app:calculation}.

%To provide lightweight and interpretable cues for routing, HTE computes a compact set of token-level statistical features (e.g., trend strength, variation, and autocorrelation) for each patch \cite{qiu2024tfb, li2024foundts}. 
%These features offer coarse summaries of local temporal behavior and are used as auxiliary signals for expert routing. Incorporating richer or higher-order statistics is possible, but our design prioritizes simplicity and robustness, and we empirically find the proposed features sufficient for effective routing.

%Given a univariate patch $s_i \in \mathbb{R}^S$, we compute three descriptors: trend strength ($c_1$), local variation ($c_2$), and autocorrelation coefficient ($c_3$).
%These features form a quantification vector $c_i = [c_1, c_2, c_3] \in \mathbb{R}^3$, which provides a coarse description of the local temporal behavior of $s_i$ and serves as the input to the expert routing mechanism. The specific calculation formulas are provided in the Appendix~\ref{sec_app:calculation}.

\textbf{Pattern-Adaptive Routing.}
Inspired by Variational Autoencoder-style stochastic modeling \cite{kingma2013auto}, we introduce latent uncertainty into the expert selection process by encoding both the input patch $s_i$ and its complexity-related statistical cues $c_i$ into latent scores. Specifically, we compute:
\begin{align}
\tilde{z}_i &= \text{ReLU}(s_i W_0^t) W_1^t, \\
\tilde{c}_i &= \text{ReLU}(c_i W_0^c) W_1^c,
\end{align}
where $W_0^t \in \mathbb{R}^{S \times d}$, $W_0^c \in \mathbb{R}^{3 \times d}$, and $W_1^t, W_1^c \in \mathbb{R}^{d \times K}$ for $K$ experts. 

We inject Gaussian noise $\epsilon_i \sim \mathcal{N}(0,1)$ and compute routing logits:
\begin{align}
z_i &= \tilde{z}_i + \epsilon_i \cdot \text{Softplus}(\tilde{c}_i), \\
h_i &= z_i W^H,
\end{align}
where $W^H \in \mathbb{R}^{K \times K}$ is a projection matrix that maps the latent vector to the expert scoring space, and $h, \epsilon \in \mathbb{R}^K$.
To promote sparsity, we retain the top-$k$ entries in $h_i$ before applying softmax:
\begin{align}
G(s_i) &= \text{Softmax}(\text{KeepTopk}(h_i, k)), \\
\text{KeepTopk}(h_i, k)_j &= 
\begin{cases} 
h_{i,j}, & \text{if } j \in \text{Topk}(h_i), \\ 
-\infty, & \text{otherwise}.
\end{cases}
\end{align}

\textbf{Heterogeneous Pattern Extractors.}
Unlike previous methods that adopt a unified architecture for all time segments \cite{nie2023time, liu2024itransformer} or apply homogeneous experts uniformly across patches \cite{sun2025learning, qiu2025duet}, we recognize that time series often exhibit diverse temporal patterns, such as trends, local fluctuations, and long-range dependencies, which motivate a heterogeneous modeling strategy.
To this end, we design a lightweight expert pool comprising three complementary branches that provide diverse temporal modeling capacities, enabling the framework to adapt to heterogeneous temporal dynamics and improve robustness across varied forecasting scenarios:

\begin{itemize}
\item Linear Expert for modeling trend-like patterns:
\begin{align}
\mathbf{e}_i^{\text{Linear}} = s_i \cdot \mathbf{W}_{\text{Linear}}.
\end{align}

\item CNN Expert for capturing local dependencies:
\begin{align}
\mathbf{e}_i^{\text{CNN}} = \mathbf{W}_{\text{proj}} \cdot (\text{Conv}_2(\text{ReLU}(\text{Conv}_1(s_i)))).
\end{align}

\item LSTM Expert for modeling long-term memory:
\begin{align}
\mathbf{e}_i^{\text{LSTM}} = \mathbf{W}_{\text{proj}} \cdot \text{LSTM}(s_i)_{[-1]},
\end{align}
\end{itemize}
where $\text{LSTM}(s_i)_{[-1]}$ denotes the hidden state of the last time step given input $s_i$.

Let $\mathbf{e}_i^j$ denote the output of the $j$-th expert. The final representation for patch $s_i$ is computed as a weighted aggregation over all expert outputs:
\begin{align}
\mathbf{e}_i = \sum_{j=1}^{K} G(s_i)_j \cdot \mathbf{e}_i^j,
\end{align}
where $G(s_i) \in \mathbb{R}^K$ is the sparse gating vector produced by the pattern-adaptive routing mechanism.

\textbf{Expert Regularization.}
To prevent expert collapse and promote diverse expert usage, we incorporate a load-balancing regularization term inspired by \cite{shazeer2017outrageously}:
\begin{align}
\mathcal{L}_{\text{MoE}} = \mathcal{L}_{\text{importance}} + \mathcal{L}_{\text{load}}.
\end{align}
Here, $\mathcal{L}_{\text{importance}}$ minimizes the coefficient of variation across expert gate importance scores, while $\mathcal{L}_{\text{load}}$ penalizes imbalanced token-to-expert assignments. This regularization stabilizes training and promotes more efficient utilization of the expert capacity.

\subsection{Representation Alignment Module}
Most existing LLM-based time series forecasting approaches rely on static, global prompts shared across all tokens \cite{jin2024timellm}, which fail to capture the temporal heterogeneity inherent in multivariate time series and limit generalization to local patterns.
Furthermore, these methods typically adopt shallow alignment strategies \cite{liu2024autotimes}, resulting in representations that are misaligned with the architecture of LLMs and fail to fully exploit their reasoning capabilities.

To address these limitations, we propose the Representation Alignment Module (RAM), which performs fine-grained token-level alignment between temporal features and their corresponding LLM-compatible representations via contrastive learning. 
By generating token-adaptive prompts and embedding both modalities into a shared latent space, RAM enables the LLM to operate on temporally grounded representations while preserving fine-grained numerical information.

\textbf{Token-Adaptive Prompt.}
Unlike language tokens that follow consistent syntactic structures, time series tokens encode heterogeneous temporal characteristics. 
Applying a uniform prompt across such tokens can obscure informative variations.
Inspired by recent advances in visual prompting \cite{liu2025token}, we extend the idea of differentiated prompts to time series.

We construct token-adaptive prompts using interpretable statistical descriptors for each token, with each prompt integrating three aspects:
(1) expert routing hints, 
(2) patch-wise temporal context, 
and (3) complexity-related features. 
Motivated by the attention analysis in \cite{liu2025timecma}, we place numerical features at the end of the prompt to guide the LLM’s focus toward informative value tokens.
These elements are tokenized using the LLM tokenizer to yield prompt embeddings ${p_i}$.

\newtcolorbox{promptbox}[2][]{
  colback=gray!10,      
  colframe=black!75!black,    
  colbacktitle=black,         
  coltitle=white,             
  title=#2,                   
  boxrule=1pt,                
  arc=1mm,                    
  enhanced,
  attach boxed title to top center={yshift=-3mm},
  boxed title style={height=6mm},  
  left=3mm,
  right=3mm,
  top=3.25mm,
  bottom=1.75mm,
  width=\linewidth,
  % center,
  #1
}

\begin{promptbox}{\textbf{Token-Adaptive Prompt}}
[Expert Routing Hint]\\
The available expert types are: Linear, CNN, LSTM.\\
                    
[Patch Time Context]\\
This patch consists of \textcolor{blue!75!black}{$\left \langle \text{token\_len} \right \rangle$} time steps, from \textcolor{blue!75!black}{$\left \langle \text{patch\_start} \right \rangle$} to \textcolor{blue!75!black}{$\left \langle \text{patch\_end} \right \rangle$}. \\
It is part of a longer input window, which spans from \textcolor{blue!75!black}{$\left \langle \text{x\_start} \right \rangle$} to \textcolor{blue!75!black}{$\left \langle \text{x\_end} \right \rangle$} and contains \textcolor{blue!75!black}{$\left \langle \text{seq\_len} \right \rangle$} time steps. \\

[Complexity-related Features]\\
Trend Strength: \textcolor{blue!75!black}{$\left \langle c_1 \right \rangle$}. \\
Local Variation: \textcolor{blue!75!black}{$\left \langle c_2 \right \rangle$}. \\
Temporal Dependency: \textcolor{blue!75!black}{$\left \langle c_3 \right \rangle$}. \\
\vspace{-10pt}
\end{promptbox}

\textbf{Representation Alignment.}
To bridge the modality gap between temporal signals and language representations, we design a contrastive alignment mechanism that injects prompt-conditioned contextual signals into temporal features at the token level.
%For each token $i$, we align its temporal feature $e_i$ with its associated prompt embedding $p_i$ via a contrastive objective:
For each token $i$, we treat $(e_i, p_i)$ as a positive pair and use the prompt embeddings $\{p_j\}_{j\ne i}$ in the same mini-batch as negative samples. The alignment objective is formulated as a standard InfoNCE loss:
\begin{align}
\mathcal{L}_{\mathrm{align}} = - \frac{1}{N} \sum^{N}_{i=1} log \frac{exp( \left \langle e_i, p_i \right \rangle / \tau)}{\sum^{N}_{j=1} exp( \left \langle e_i, p_j \right \rangle / \tau)},
\end{align}
where $\langle \cdot, \cdot \rangle$ denotes cosine similarity, $\tau$ is a temperature parameter, and all vectors are $\ell_2$-normalized.

This alignment encourages temporal features to reside in a shared representation space with their corresponding prompts, enabling consistent interaction with the pretrained LLM while preserving fine-grained temporal information.

\subsection{LLM Forecasting Head}
By aligning temporal features with LLM-compatible representations, we enable the LLM to operate on time series in a representation space that is compatible with its pretrained architecture. 
The aligned features ${e_i}$ are first passed through a frozen pretrained LLM for deep contextual modeling, after which a lightweight decoder projects the resulting representations into future predictions:
\begin{align}
\hat{Y} = \text{MLP}(\text{LLM}(e)).
\end{align}
Our autoregressive decoding allows flexible forecasting without retraining for different horizons, fully utilizing LLMs’ inherent capacity for multi-step generation \cite{liu2024autotimes}.

The final training objective jointly optimizes forecasting accuracy, expert utilization, and cross-modal representation alignment:
\begin{align}
\mathcal{L} = \mathcal{L}_{\mathrm{MSE}} + \alpha \mathcal{L}_{\mathrm{MoE}} + \beta \mathcal{L}_{\mathrm{align}}.
\end{align}
This formulation enables accurate and generalizable forecasts while maintaining an efficient decoding pipeline.

\begin{table*}[t]
\caption{Multivariate forecasting (672-pred-{96, 192, 336, 720}) results under the one-for-all setting. Following \cite{liu2024autotimes}, a single model is trained on a 96-step prediction horizon and evaluated on all horizons using rolling forecasting. The best results are in \textbf{bold}, and the second-best are \textit{underlined}. IMP denotes the average MSE and MAE reduction of TALON over each baseline across seven datasets.}
\label{tab:Rolling Forecast Results}
\begin{center}
\resizebox{1.0\linewidth}{!}{
\large
\begin{tabular}{cc|cccccccccccc|cccccccccccccc}
\toprule
\multicolumn{2}{c|}{}                                    & \multicolumn{12}{c|}{LLM-based methods}                                                                                                                                                                                                                                                                                     & \multicolumn{14}{c}{Deep learning forecasting methods}                                                                                                                                                                                                        \\ \cmidrule{3-28} 
\multicolumn{2}{c|}{}                                    & \multicolumn{2}{c|}{TALON}                                                                 & \multicolumn{2}{c|}{LangTime}                        & \multicolumn{2}{c|}{CALF}                            & \multicolumn{2}{c|}{AutoTimes}                 & \multicolumn{2}{c|}{TimeLLM}       & \multicolumn{2}{c|}{FPT}     & \multicolumn{2}{c|}{SimpleTM}                  & \multicolumn{2}{c|}{Timer\_XL}                 & \multicolumn{2}{c|}{PPGF}            & \multicolumn{2}{c|}{TimeMixer}           & \multicolumn{2}{c|}{iTransformer}  & \multicolumn{2}{c|}{PatchTST}               & \multicolumn{2}{c}{TimesNet} \\
\bottomrule
\end{tabular}}
\end{center}
\end{table*}

\subsection{Inference Pipeline}
As shown in Figure~\ref{fig:model}, the inference process is streamlined and fully prompt-free.
The input time series is first segmented into patches $s_1, s_2, \dots, s_N$, and each patch is processed by the Heterogeneous Temporal Encoder.
A gating mechanism aggregates outputs from the top-$k$ experts, yielding LLM-compatible and context-aware features $e_1, e_2, \dots, e_N$, which are then passed through a frozen pretrained LLM to perform autoregressive forecasting.

The token-adaptive prompts are only used during training to provide semantic anchors for contrastive representation alignment. After training, the model has already learned to map time-series patches into an LLM-compatible representation space, so explicit prompts are no longer needed at inference. In this sense, prompt-guided alignment is used to shape the feature space during training, while prompt-free inference directly exploits the learned aligned temporal representations.

By eliminating the need for textual prompts and representation alignment during inference, our framework supports efficient, pattern-aware forecasting with minimal computational overhead. This design enables faster inference and enhanced deployment flexibility, while retaining the representational benefits of heterogeneous expert modeling.

\section{Experiment}
\label{sec:Experiment}
\subsection{Data and Experiment Setting}

\textbf{Dataset.}
We evaluate the long-term forecasting performance across seven widely-used time series benchmarks, including ETT datasets (ETTh1, ETTh2, ETTm1, ETTm2), Weather, Electricity, and Traffic. These datasets are standard benchmarks in the long-term forecasting literature \cite{liu2024autotimes}. Detailed descriptions are provided in Appendix~\textcolor{mypink}{A}.%Appendix~\ref{app:dataset}.

\begin{table*}[t]
\caption{Multivariate forecasting (672-pred-\{96, 192, 336, 720\}) results under the one-for-one setting. A separate model is trained and evaluated for each prediction horizon. The best results are in \textbf{bold}, and the second-best are \textit{underlined}.}
\label{tab:Forecast Results}
\begin{center}
\resizebox{1.0\linewidth}{!}{
\large
\begin{tabular}{cc|cc|cccccccccccccccccccccccc}
\toprule
\multicolumn{2}{c|}{\multirow{3}{*}{Models}}            & \multicolumn{2}{c|}{One-for-all} & \multicolumn{24}{c}{Trained   respectively on specific lookback / prediction length}                                                                                                                                                                                                                                                                                                                                                                                                                                                        \\ \cmidrule{3-28} 
\multicolumn{2}{c|}{}                                   & \multicolumn{2}{c|}{TALON}       & \multicolumn{2}{c|}{LangTime}                     & \multicolumn{2}{c|}{CALF}                         & \multicolumn{2}{c|}{AutoTimes}              & \multicolumn{2}{c|}{TimeLLM}       & \multicolumn{2}{c|}{FPT}                          & \multicolumn{2}{c|}{SimpleTM}                        & \multicolumn{2}{c|}{Timer\_XL}                  & \multicolumn{2}{c|}{PPGF}                       & \multicolumn{2}{c|}{TimeMixer}                    & \multicolumn{2}{c|}{iTransformer}           & \multicolumn{2}{c|}{PatchTST}                     & \multicolumn{2}{c}{TimesNet} \\
\bottomrule
\end{tabular}}
\end{center}
\end{table*}

\textbf{Baselines and Evaluation.}
We compare TALON against state-of-the-art baselines from two categories:
(1) LLM-based forecasting methods: LangTime \cite{niu2025langtime}, CALF \cite{liu2025calf}, AutoTimes \cite{liu2024autotimes}, TimeLLM \cite{jin2024timellm}, and FPT \cite{zhou2023one};
(2) Deep learning-based forecasting models: SimpleTM \cite{chen2025simpletm}, Timer\_XL \cite{liu2025timerxl}, PPGF \cite{sun2025ppgf}, TimeMixer \cite{wang2024timemixer}, iTransformer \cite{liu2024itransformer}, PatchTST \cite{nie2023time}, and TimesNet \cite{wu2023timesnet}.
All baselines are reproduced using their official implementations to ensure fair comparison.

\textbf{Implementation Details.}
Following the common setup in \cite{liu2024autotimes}, we fix the input lookback window size to $L = 672$ for all experiments and use pre-trained GPT2 based model \cite{radford2019language} with the first 6 Transformer layers as our backbone. To ensure fair comparisons, we rerun all baselines. All the experiments are conducted using PyTorch \cite{NEURIPS2019_bdbca288} on NVIDIA A100 GPUs.

\subsection{Time Series Forecasting}
\textbf{Setups.} We consider two evaluation protocols to assess the forecasting performance of our model:
(1) To evaluate the generalization capability of one-for-all forecasting, we adopt the rolling forecast setting \cite{liu2024autotimes, liu2025timerxl}, where a single model is trained on a 96-step prediction horizon and then directly applied to all other horizons. During inference, the predicted values are recursively fed into the lookback window to generate subsequent predictions.
(2) For the conventional one-for-one setting, we follow the standard multivariate evaluation protocol adopted by TimesNet \cite{wu2023timesnet}, where a separate model is trained and evaluated for each prediction horizon.

\textbf{Results.}
The average forecasting results are reported in Table~\ref{tab:Rolling Forecast Results} and Table~\ref{tab:Forecast Results}.
In the one-for-all setting (Table~\ref{tab:Rolling Forecast Results}), TALON consistently achieves the lowest MSE across all seven datasets, with an average improvement of up to 10\% over state-of-the-art deep forecasters and 12\% over recent LLM-based methods.
In the conventional one-for-one setting (Table~\ref{tab:Forecast Results}), it further achieves state-of-the-art performance with up to 20\% MSE reduction.
These results highlight TALON’s strong generalization capability and its effectiveness in modeling heterogeneous and evolving temporal patterns.

\begin{table}[t]
\caption{Comparison between TALON and MoE-based methods.}
\label{tab:moe_comparison}
\begin{center}
\resizebox{1.0\linewidth}{!}{
\large
\begin{tabular}{cc|cc|cc|cc|cc|cc}
\toprule
\multicolumn{2}{c|}{\multirow{2}{*}{Models}}      & \multicolumn{2}{c|}{TALON}      & \multicolumn{2}{c|}{FreqMoE} & \multicolumn{2}{c|}{MoFE-time}  & \multicolumn{2}{c|}{TimeMoE} & \multicolumn{2}{c}{TFPS}        \\ \cmidrule{3-12} 
\multicolumn{2}{c|}{}                             & \multicolumn{2}{c|}{(Ours)}     & \multicolumn{2}{c|}{(2025) \cite{liu2025freqmoe}}    & \multicolumn{2}{c|}{(2025) \cite{liu2025mofe}}      & \multicolumn{2}{c|}{(2025) \cite{shi2025timemoe}}    & \multicolumn{2}{c}{(2025) \cite{sun2025learning}} \\ \midrule
\multicolumn{2}{c|}{Metric}                       & MSE            & MAE            & MSE         & MAE            & MSE            & MAE            & MSE           & MAE          & MSE            & MAE            \\ \midrule
\multicolumn{1}{c|}{\multirow{4}{*}{ETTh1}} & 96  & {\underline{0.351}}    & 0.392          & 0.371       & {\underline{0.388}}    & \textbf{0.337} & \textbf{0.380} & 0.360         & 0.396        & 0.398          & 0.413          \\
\multicolumn{1}{c|}{}                       & 192 & \textbf{0.381} & 0.415          & 0.426       & 0.422          & {\underline{0.381}}    & \textbf{0.411} & 0.386         & {\underline{0.413}}  & 0.423          & 0.423          \\
\multicolumn{1}{c|}{}                       & 336 & \textbf{0.398} & \textbf{0.427} & 0.475       & 0.447          & 0.414          & 0.436          & {\underline{0.407}}   & {\underline{0.433}}  & 0.484          & 0.461          \\
\multicolumn{1}{c|}{}                       & 720 & \textbf{0.414} & \textbf{0.446} & 0.488       & {\underline{0.459}}    & {\underline{0.453}}    & 0.466          & 0.457         & 0.476        & 0.488          & 0.476          \\ \midrule
\rowcolor{gray!30}   
\multicolumn{2}{c|}{Avg.}                         & \textbf{0.386} & \textbf{0.420} & 0.440       & 0.429          & {\underline{0.396}}    & {\underline{0.423}}    & 0.402         & 0.429        & 0.448          & 0.443          \\ \midrule
\multicolumn{1}{c|}{\multirow{4}{*}{ETTh2}} & 96  & \underline{0.302} & {\underline{0.349}}    & {\textbf{0.287}} & \textbf{0.337} & 0.307          & 0.352          & 0.352         & 0.388        & 0.313          & 0.355          \\
\multicolumn{1}{c|}{}                       & 192 & \textbf{0.355} & {\underline{0.388}}    & {\underline{0.361}} & \textbf{0.386} & 0.389          & 0.418          & 0.425         & 0.434        & 0.405          & 0.410          \\
\multicolumn{1}{c|}{}                       & 336 & \textbf{0.371} & \textbf{0.406} & 0.407       & 0.423          & 0.514          & 0.480          & 0.526         & 0.485        & {\underline{0.392}}    & {\underline{0.415}}    \\
\multicolumn{1}{c|}{}                       & 720 & \textbf{0.393}          & \underline{0.435}          & 0.414 & 0.438    & 0.543          & 0.505          & 0.585         & 0.526        & \underline{0.410} & \textbf{0.433} \\ \midrule
\rowcolor{gray!30}   
\multicolumn{2}{c|}{Avg.}                         & \textbf{0.355} & {\underline{0.401}}    & {\underline{0.367}} & \textbf{0.396} & 0.438          & 0.439          & 0.472         & 0.458        & 0.380          & 0.403          \\ \midrule
\multicolumn{1}{c|}{\multirow{4}{*}{ETTm1}} & 96  & \textbf{0.278} & \textbf{0.339} & 0.314       & 0.356          & {\underline{0.294}}    & {\underline{0.352}}    & 0.319         & 0.373        & 0.327          & 0.367          \\
\multicolumn{1}{c|}{}                       & 192 & \textbf{0.324} & \textbf{0.367} & 0.356       & {\underline{0.380}}    & {\underline{0.333}}    & 0.381          & 0.359         & 0.401        & 0.374          & 0.395          \\
\multicolumn{1}{c|}{}                       & 336 & \textbf{0.358} & \textbf{0.388} & {\underline{0.385}} & {\underline{0.404}}    & 0.400          & 0.433          & 0.404         & 0.433        & 0.401          & 0.408          \\
\multicolumn{1}{c|}{}                       & 720 & \textbf{0.418} & \textbf{0.424} & {\underline{0.446}} & {\underline{0.445}}    & 0.536          & 0.514          & 0.545         & 0.501        & 0.479          & 0.456          \\ \midrule
\rowcolor{gray!30}   
\multicolumn{2}{c|}{Avg.}                         & \textbf{0.345} & \textbf{0.380} & {\underline{0.375}} & {\underline{0.396}}    & 0.391          & 0.420          & 0.407         & 0.427        & 0.395          & 0.407          \\ \midrule
\multicolumn{1}{c|}{\multirow{4}{*}{ETTm2}} & 96  & {\underline{0.173}}    & {\underline{0.260}}    & 0.173       & 0.266          & 0.189          & 0.278          & 0.258         & 0.320        & \textbf{0.170} & \textbf{0.255} \\
\multicolumn{1}{c|}{}                       & 192 & \textbf{0.230} & {\underline{0.299}}    & {\underline{0.235}} & 0.310          & 0.249          & 0.327          & 0.270         & 0.338        & 0.235          & \textbf{0.296} \\
\multicolumn{1}{c|}{}                       & 336 & \textbf{0.282} & \textbf{0.333} & {\underline{0.290}} & 0.350          & 0.294          & 0.356          & 0.365         & 0.405        & 0.297          & {\underline{0.335}}    \\
\multicolumn{1}{c|}{}                       & 720 & \textbf{0.362} & \textbf{0.383} & 0.385       & 0.424          & {\underline{0.381}}    & 0.425          & 0.403         & 0.445        & 0.401          & {\underline{0.397}}    \\ \midrule
\rowcolor{gray!30}   
\multicolumn{2}{c|}{Avg.}                         & \textbf{0.262} & \textbf{0.319} & {\underline{0.271}} & 0.338          & 0.278          & 0.347          & 0.324         & 0.377        & 0.276          & {\underline{0.321}}    \\ \midrule
\rowcolor{gray!30}  
\multicolumn{2}{c|}{IMP.}              & -- --              & -- --              & 7\%           & 3\%              & {\underline{10\%}}              & {\underline{7\%}}             & \textbf{16\%}             & \textbf{11\%}            & 10\%        & 4\%             \\ \bottomrule
\end{tabular}}
\end{center}
\end{table}

\subsection{Compared with MoE-based Methods}
As shown in Table~\ref{tab:moe_comparison}, TALON achieves the best MSE scores across all datasets, with an average improvement of 7\% to 16\% over existing methods, where TimeMoE results are taken from MoFE-Time due to the lack of per-dataset reporting in the original paper, while other baselines follow their original reports.
These consistent gains highlight the advantage of heterogeneous expert modeling: by incorporating diverse inductive biases, TALON adapts to evolving temporal dynamics and heterogeneous patterns across segments, leading to more reliable and robust forecasts. 
This demonstrates the importance of architectural diversity in enhancing model generalization and handling non-stationary dynamics across time series segments.

\subsection{Zero-shot Forecasting}
\textbf{Setups.}
LLMs have exhibited remarkable zero-shot generalization capabilities across various domains \cite{brown2020language}.
To assess whether TALON inherits this ability, we adopt the widely used zero-shot forecasting protocol \cite{jin2024timellm, liu2024autotimes}, where a model is trained on a source domain and directly evaluated on an unseen target domain without any fine-tuning. 
Following this setting, we use the ETT benchmark family and conduct evaluations across multiple cross-domain scenarios, including both resolution shifts and domain shifts among ETT variants. 
As in the full-shot experiments, we adopt the trained model under the one-for-all forecasting protocol for evaluation.

\textbf{Results.}
The zero-shot forecasting results are summarized in Table~\ref{tab:Zero-shot Results}.
TALON shows competitive MSE performance in 4 tasks relative to the compared methods. Specifically, it achieves 8\%$\sim$20\% relative MSE improvement, demonstrating robust generalization across diverse transfer scenarios, including both resolution-level shifts and cross-domain adaptations.
These results validate the effectiveness of TALON in capturing local temporal structures and leveraging LLM-compatible representation alignment for strong transferability.

\begin{table}[t]
\caption{Zero-shot forecasting result.}
\label{tab:Zero-shot Results}
\begin{center}
\resizebox{0.98\linewidth}{!}{
\large
\begin{tabular}{ccc|cc|cc|cc|cc}
\toprule
\multicolumn{3}{c|}{}                                                                              & \multicolumn{2}{c|}{TALON}                                                      & \multicolumn{2}{c|}{LangTime}                                          & \multicolumn{2}{c|}{AutoTimes}                                               & \multicolumn{2}{c}{Timer\_XL}                                              \\
\multicolumn{3}{c|}{\multirow{-2}{*}{Model}}                                                       & \multicolumn{2}{c|}{(Ours)}                                                     & \multicolumn{2}{c|}{(2025) \cite{niu2025langtime}}                                              & \multicolumn{2}{c|}{(2024) \cite{liu2024autotimes}}                                                    & \multicolumn{2}{c}{(2025) \cite{liu2025timerxl}}                                                  \\ \midrule
\multicolumn{3}{c|}{Metric}                                                                        & MSE                                    & MAE                                    & MSE                           & MAE                                    & MSE                                    & MAE                                 & MSE                                 & MAE                                 \\ \midrule
\multicolumn{1}{c|}{}                         & \multicolumn{1}{c|}{}                        & 96  & \textbf{0.290}                         & \textbf{0.348}                         & 0.338                         & 0.373                                  & {\underline{0.294}}                            & {\underline{0.352}}                         & 0.305                               & 0.358                               \\
\bottomrule
\end{tabular}}
\end{center}
\end{table}

\subsection{Model Analysis}
\label{sec:Model Analysis}
\begin{figure}[t]
    \centering
    \includegraphics[width=1.0\linewidth]{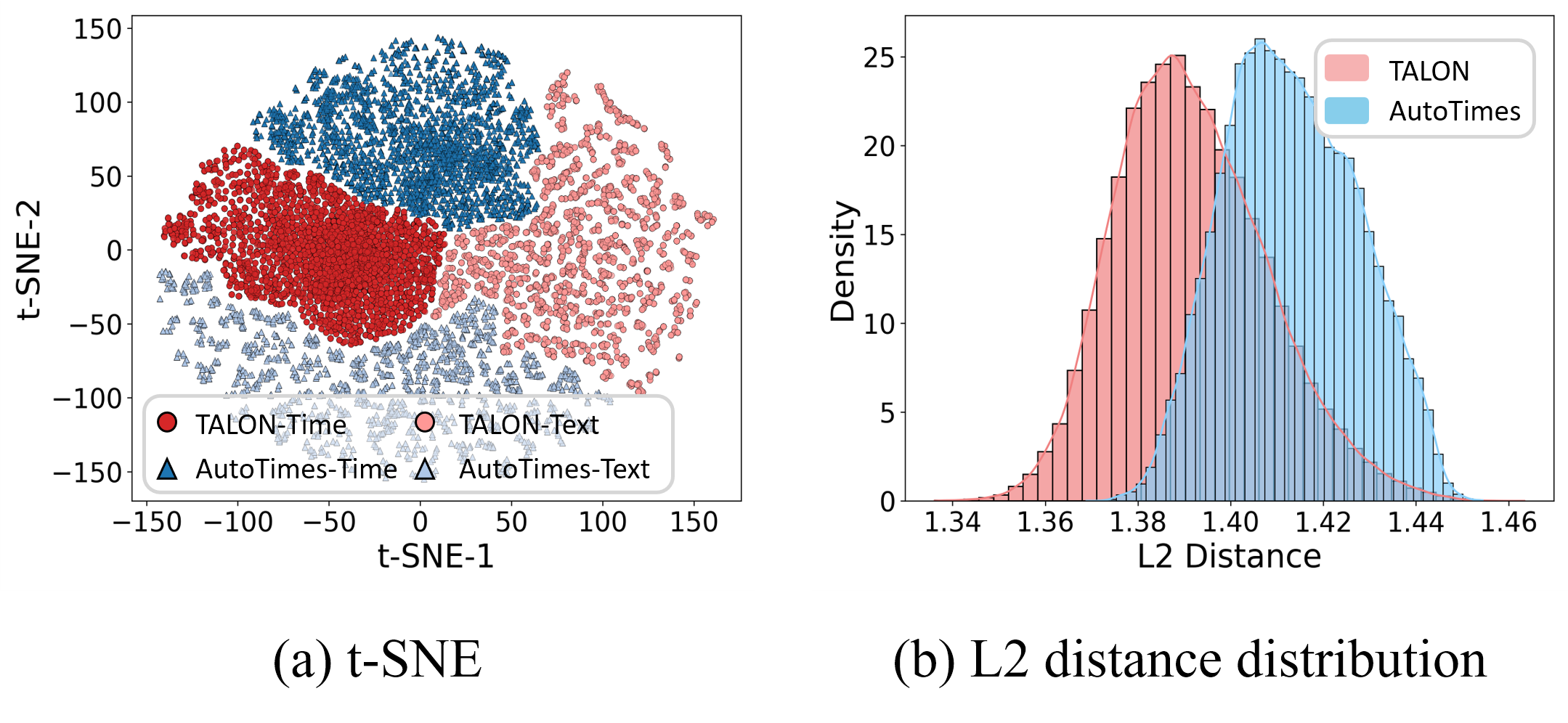}
    \caption{Visualization of time–text alignment for TALON and AutoTimes on ETTh1-96.}
    \label{fig:alignment}
\end{figure}

\textbf{Cross-Modal Embedding Alignment Analysis.}
To evaluate the quality of cross-modal alignment, we analyze both the spatial structure and the quantitative similarity between time-series and textual embeddings. As shown in Figure~\ref{fig:alignment}~(a), the t-SNE visualization shows that TALON’s temporal and prompt-derived embeddings form more compact clusters, indicating stronger cross-modal representation consistency. In contrast, AutoTimes exhibits a more scattered distribution, suggesting weaker alignment between modalities.
We also compute the L2 distance between aligned time-text embedding pairs across the test set. As shown in Figure~\ref{fig:alignment}~(b), TALON achieves a significantly smaller mean distance than AutoTimes, confirming its stronger cross-modal correspondence.

\textbf{Ablation Studies.}
We conduct ablation studies to evaluate the contributions of TALON’s key components. As shown in Table~\ref{tab:Ablation}, removing the full HTE module (w/o HTE) increases average MSE by 8.6\%, while disabling only the routing mechanism (w/o HTE\_R) leads to an 8.1\% increase, highlighting the value of expert specialization and routing.
Disabling RAM (w/o RAM) results in a 7.2\% increase in MSE, demonstrating its benefit in aligning temporal and textual representations.
Replacing our token-adaptive prompt with a static TimeLLM-style prompt (w/o Prompt) leads to 5.6\% degradation, validating the design of context-aware prompt construction.
Removing the LLM (w/o LLM) causes the most significant drop, with a 33.9\% increase in MSE, indicating the essential role of the LLM as the underlying reasoning backbone.
At the same time, the consistent degradation caused by removing HTE, routing, RAM, or the token-adaptive prompt shows that TALON’s gains do not stem from LLM capacity alone, but from the synergy between the pretrained LLM and the proposed temporal adaptation modules.
These results confirm that each module plays a meaningful role in TALON, and their combination produces a synergistic effect in modeling complex, heterogeneous temporal dynamics.

\begin{table}[t]
\caption{Performance of ablation studies.}
\label{tab:Ablation}
\begin{center}
\resizebox{1.0\linewidth}{!}{
\large
\begin{tabular}{c|cc|cc|cc|cc}
\toprule
Models & \multicolumn{2}{c|}{ETTh1}      & \multicolumn{2}{c|}{ETTh2}      & \multicolumn{2}{c|}{ETTm1}      & \multicolumn{2}{c}{ETTm2}       \\ \midrule 
Metric                       & MSE            & MAE            & MSE            & MAE            & MSE            & MAE            & MSE            & MAE            \\ \midrule
\rowcolor{gray!30}  
TALON             & \textbf{0.386} & \textbf{0.420} & \textbf{0.355} & \textbf{0.395} & \textbf{0.345} & \textbf{0.380} & \textbf{0.259} & \textbf{0.319} \\
w/o HTE                & 0.403          & 0.427          & 0.365          & 0.405          & 0.347          & 0.380          & 0.267          & 0.325          \\
w/o HTE\_R              & 0.403          & 0.426          & 0.360          & 0.400          & 0.349          & 0.382          & 0.268          & 0.323          \\
w/o RAM                & 0.393          & 0.422          & 0.367          & 0.408          & 0.350          & 0.382          & 0.266          & 0.319          \\
w/o Prompt             & 0.389          & 0.419          & 0.363          & 0.406          & 0.352          & 0.383          & 0.266          & 0.322          \\
w/o LLM                & 0.418          & 0.435          & 0.386          & 0.434          & 0.396          & 0.411          & 0.280          & 0.333
   \\ \bottomrule
\end{tabular}}
\end{center}
\end{table}

\begin{figure}[t]
    \centering
    \begin{minipage}{0.48\linewidth}
        \centering
        \includegraphics[width=\linewidth]{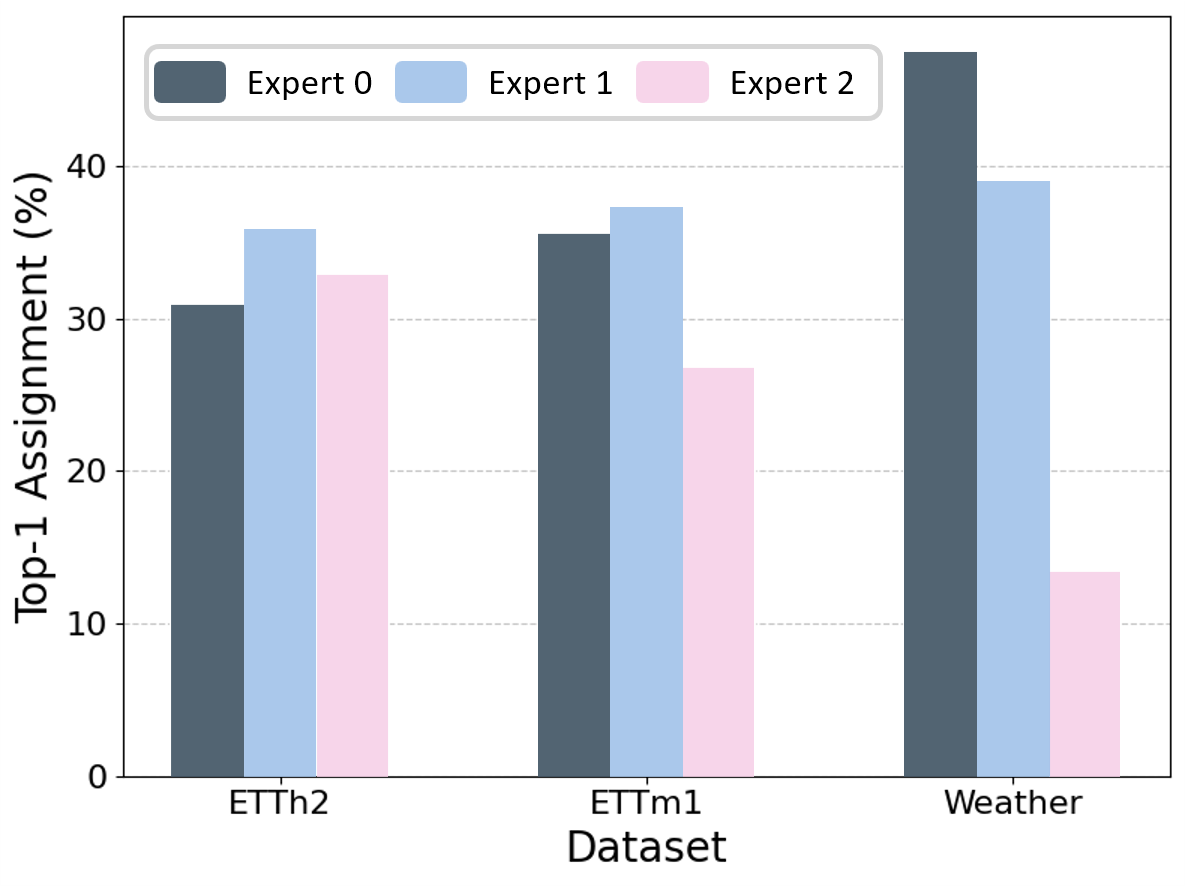}
        \caption{Analysis of expert assignment distributions.}
        \label{fig:expert_dist}
    \end{minipage}
    \hfill
    \begin{minipage}{0.5\linewidth}
        \centering
        \includegraphics[width=\linewidth]{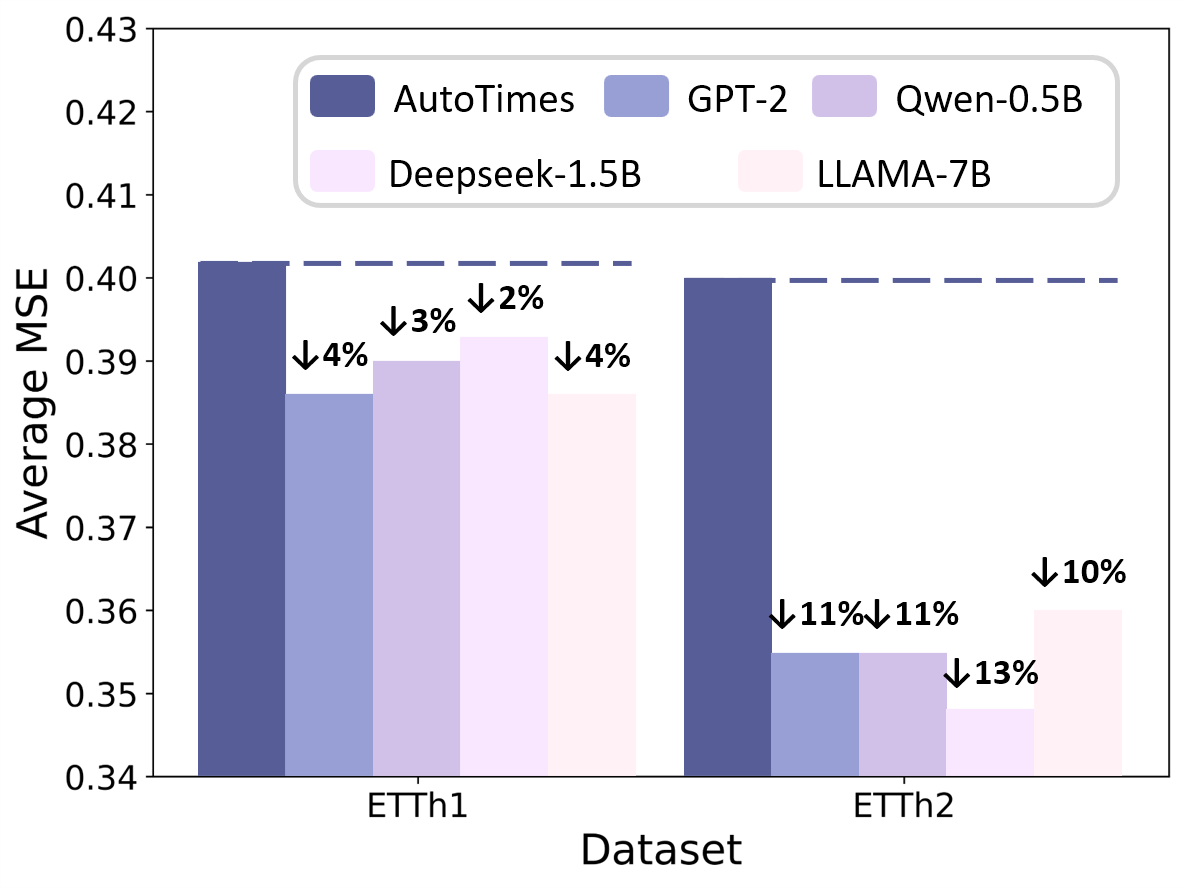}
        \caption{TALON generalization with different LLM backbones.}
        \label{fig:generality}
    \end{minipage}
\end{figure}

\textbf{Expert Assign.}
Figure~\ref{fig:expert_dist} illustrates the expert assignment distributions of TALON across ETTh2, ETTm1, and Weather. Each bar indicates the percentage of input segments that are most confidently routed to a given expert.
We observe that the expert utilization patterns vary significantly across datasets. For example, the Weather dataset shows a strong preference for Expert 0, whereas ETTh2 and ETTm1 exhibit more balanced and diverse assignments, indicating higher pattern heterogeneity \cite{sun2025learning}.
This variation highlights TALON’s ability to adaptively route segments to specialized experts based on underlying pattern characteristics, validating the effectiveness of its pattern-aware routing mechanism.

\textbf{Generality.}
Previous LLM4TS approaches \cite{zhou2023one, jin2024timellm} typically target specific language models. In contrast, TALON is designed to be compatible with any decoder-only LLM. We evaluate this generality by replacing the default GPT-2 backbone with representative alternatives: Qwen \cite{team2024qwen2}, Deepseek \cite{liu2024deepseek}, and LLaMA \cite{touvron2023llama}.
We choose AutoTimes as the baseline, as it exhibits the smallest relative performance improvement (5\% in MSE) under TALON in Table~\ref{tab:Rolling Forecast Results}.
As shown in Figure~\ref{fig:generality}, TALON consistently outperforms AutoTimes across all datasets and LLMs, with relative MSE reductions annotated on each bar. These results confirm that our framework is reliably enhances forecasting performance regardless of the underlying LLM. 
%Interestingly, forecasting performance does not monotonically scale with model size: smaller models such as GPT-2 sometimes outperform larger ones, suggesting that pretraining corpus, architectural choices, and tokenization strategies are critical factors beyond parameter count. While a systematic study of LLM scale is left for future work, our results demonstrate that TALON delivers consistent improvements across diverse backbones, highlighting its general applicability.
Interestingly, forecasting performance does not monotonically scale with model size: smaller models such as GPT-2 sometimes outperform larger ones, suggesting that pretraining corpus, architectural choices, and tokenization strategies are critical factors beyond parameter count \cite{gruver2023large, hua2026diversified}. While a systematic study of LLM scale is left for future work, our results demonstrate that TALON delivers consistent improvements across diverse backbones, highlighting its general applicability.

\textbf{Analysis of HTE.}
Table~\ref{tab:HTE} validates the effectiveness of the HTE design. The fully heterogeneous setup consistently achieves the best performance across all datasets. 
In contrast, removing any single expert type leads to notable performance degradation (8.1\%, 7.9\%, and 5.8\%, respectively).
These results underscore the complementary nature of distinct temporal modeling perspectives. Their integration enables the model to adapt to diverse temporal patterns within multivariate time series, thereby enhancing generalization across different forecasting scenarios.

\begin{table}[t]
\caption{Effectiveness of heterogeneous experts in HTE.}
\label{tab:HTE}
\begin{center}
\resizebox{1.0\linewidth}{!}{
\large
\begin{tabular}{ccc|cc|cc|cc|cc}
\toprule
\multicolumn{3}{c|}{Heterogeneous Experts} & \multicolumn{2}{c|}{ETTh1}       & \multicolumn{2}{c|}{ETTh2}       & \multicolumn{2}{c|}{ETTm1}       & \multicolumn{2}{c}{ETTm2}       \\ \midrule 
Linear      & CNN      & LSTM      & MSE            & MAE            & MSE            & MAE            & MSE            & MAE            & MSE            & MAE            \\ \midrule
\rowcolor{gray!30}  
\checkmark           & \checkmark        & \checkmark         & \textbf{0.386} & \textbf{0.420} & \textbf{0.355} & \textbf{0.395} & \textbf{0.345} & \textbf{0.380} & \textbf{0.259} & \textbf{0.319} \\
\XSolidBrush           & \checkmark         & \checkmark         & 0.389          & 0.419          & 0.370          & 0.407          & 0.354          & 0.385          & 0.266          & 0.322          \\
\checkmark           & \XSolidBrush        & \checkmark         & 0.401          & 0.426          & 0.360          & 0.400          & 0.353          & 0.386          & 0.265          & 0.320          \\
\checkmark           & \checkmark        & \XSolidBrush         & 0.393          & 0.422          & 0.363          & 0.405          & 0.350          & 0.382          & 0.263          & 0.320          \\ \bottomrule
\end{tabular}}
\end{center}
\end{table}

\begin{figure}[t]
    \centering
    \includegraphics[width=0.65\linewidth]{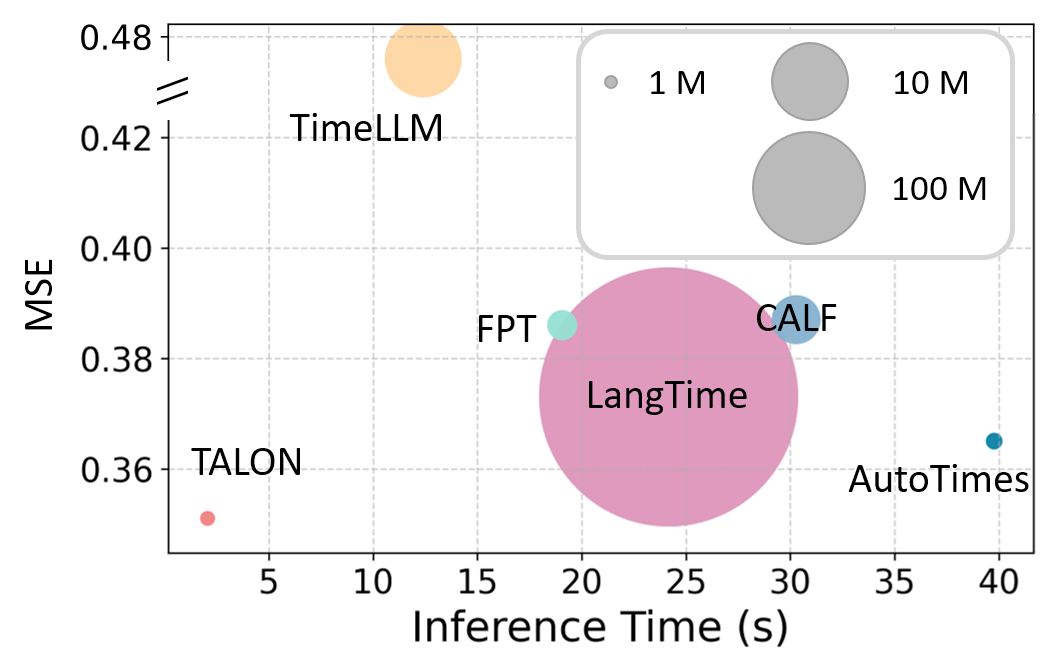}
    \caption{Efficiency comparison across LLM-based forecasters on ETTh1-96.}
    \label{fig:Efficiency}
\end{figure}

\textbf{Efficiency Analysis.} 
As shown in Figure~\ref{fig:Efficiency}, we compare TALON’s efficiency with other LLM-based models on ETTh1-96.
TALON achieves the lowest MSE while maintaining a compact model size ($\sim$ 1.7M) and fast inference ($\sim$ 2s), showing that careful architectural design can improve accuracy without increasing computational cost. 
This efficiency stems from TALON’s lightweight temporal encoder and prompt-free representation alignment, which together reduce input redundancy by removing handcrafted prompts and mitigate input complexity by preserving the temporal continuity and numerical precision of the original series.
For clarity, the reported parameter count corresponds to the trainable parameters of each method. Therefore, frozen LLM backbones are excluded from the parameter count, while LLM parameters are included for methods that require LLM fine-tuning.

%\subsection{Visualization}
%To qualitatively assess the forecasting capability of the proposed TALON model, we present representative visualizations on the ETTh1-192 dataset. 
%We compare TALON with AutoTimes~\cite{liu2024autotimes}, a strong baseline that achieves the second-best MSE performance on ETTh1-192.
%As illustrated in Figure~\ref{fig:forecast-examples}, TALON consistently produces forecasts that align more closely with the ground truth. 
%These visualizations indicate that TALON not only achieves superior quantitative performance, but also generates more realistic and faithful predictions, particularly in capturing complex patterns and fluctuations in real-world time series.
%Additional qualitative examples are provided in Appendix~\textcolor{mypink}{F}.%Appendix~\ref{app_sec:visualization}.

%\begin{figure}[t]
%    \centering
%        \includegraphics[width=0.48\linewidth]{Figure/ETTh1_TALON_192.png}\hspace{0.01\linewidth} 
%        \includegraphics[width=0.48\linewidth]{Figure/ETTh1_AutoTimes_192.png}
%        \caption{Forecasting examples across ETTh1 (672-step input, 192-step prediction).}
%    \label{fig:forecast-examples}
%\end{figure}

\section{Conclusion}
This paper presents TALON, a novel framework for time series forecasting that integrates temporal heterogeneity modeling and representation alignment within a unified foundation model architecture.
By incorporating a heterogeneous temporal encoder and a representation-aware fusion mechanism, TALON enables off-the-shelf large language models to perform pattern-aware and LLM-compatible forecasting across diverse scenarios.
Extensive experiments on multiple benchmarks demonstrate that TALON achieves state-of-the-art accuracy while maintaining high efficiency and scalability. It also generalizes well in zero-shot settings and seamlessly incorporates both numerical and textual temporal cues.
In future work, we plan to further improve pattern modeling via more adaptive and fine-grained mechanisms, and enhance domain transferability through efficient adaptation techniques.
%such as low-rank tuning.

%\clearpage
%\newpage
\appendix
\section{Implementation Details}
\label{app_sec:Implementation Details}

\subsection{Benchmark Datasets}
\label{app:dataset}
To evaluate the effectiveness and generalization ability of our proposed model, we conduct experiments on seven widely-used benchmark datasets, covering a diverse range of domains including electricity, traffic, and weather. The detailed dataset statistics are summarized in Table~\ref{app_tab:dataset}.

\begin{itemize}
\item \textbf{ETTh1 \& ETTh2:} These datasets are part of the Electricity Transformer Temperature (ETT) benchmark, which records hourly temperature readings from two electricity transformers. Each dataset contains 7 variables.

\item \textbf{ETTm1 \& ETTm2:} These are the minute-level variants of the ETT benchmark, with a finer temporal granularity of 15 minutes. Each dataset contains 7 variables.% and significantly more samples due to the higher sampling rate.

\item \textbf{Weather:} This dataset includes 21 meteorological variables, such as temperature, humidity, and wind speed, recorded every 10 minutes in 2020 at the Max Planck Biogeochemistry Institute's weather station.

\item \textbf{Electricity:} This dataset records hourly electricity consumption for 321 clients. Due to its multivariate nature and high dimensionality, it is commonly used to evaluate model scalability and performance in high-dimensional forecasting tasks.

\item \textbf{Traffic:} This dataset records hourly occupancy rates from 862 road sensors on freeways in the San Francisco Bay Area, spanning from January 2015 to December 2016. Its high dimensionality and complex temporal patterns make it a challenging benchmark for multivariate long-term forecasting.
\end{itemize}

We follow the same data processing and train-validation-test set split protocol used in TimesNet \cite{wu2023timesnet}, where the train, validation, and test datasets are strictly divided according to chronological order to ensure no data leakage. 
For long-term forecasting, we fix the context length of TALON and the lookback window of other baseline models to 672, while the prediction lengths vary among \{96, 192, 336, 720\}. Detailed settings are summarized in Table~\ref{app_tab:dataset}.

\begin{table}[!htbp]
\caption{Detailed dataset descriptions. Dim denotes the variate number. Dataset Size denotes the total number of time points in (Train, Validation, Test) split respectively. Forecast Length denotes the future time points to be predicted. Frequency denotes the sampling interval of time points.}
\label{app_tab:dataset}
\begin{center}
\resizebox{1.0\linewidth}{!}{
\large
\begin{tabular}{c|c|c|c|c|c}
\toprule
Dataset     & Dim & Forecast Length       & Dataset Size          & Frequency                 & Information \\ \midrule
ETTh1       & 7   & \{96, 192, 336, 720\} & (8545, 2881, 2881)    & 1 hour                    & Electricity \\
ETTh2       & 7   & \{96, 192, 336, 720\} & (8545, 2881, 2881)    & 1 hour                    & Electricity \\
ETTm1       & 7   & \{96, 192, 336, 720\} & (34465, 11521, 11521) & 15 min                    & Electricity \\
ETTm2       & 7   & \{96, 192, 336, 720\} & (34465, 11521, 11521) & 15 min                    & Electricity \\
Weather     & 21  & \{96, 192, 336, 720\} & (36792, 5271, 10540)  & 10 min                    & Weather     \\ 
Electricity & 321 & \{96, 192, 336, 720\} & (18317, 2633, 5261)   & 1 hour                    & Electricity \\
Traffic     & 862 & \{96, 192, 336, 720\} & (12185, 1757, 3509)   & 1 hour                    & Transportation\\ \bottomrule
\end{tabular}}
\end{center}
\end{table}
\subsection{Implementation Details}
TALON encodes statistical information in natural language form and uses a pretrained LLM (GPT2 \cite{achiam2023gpt}) to obtain prompt embeddings by extracting the final token’s representation \cite{liu2024autotimes, liu2025timecma}. 
For multivariate forecasting, prompts are constructed independently for each variable and pre-tokenized to avoid runtime overhead.

After obtaining the prompt embeddings, TALON repurposes the LLM for time series forecasting.
During training, only the parameters of the Heterogeneous Temporal Encoder and Forecast Head are updated, while the LLM remains frozen. 
At inference, TALON employs autoregressive decoding over language-aligned features to generate variable-length predictions without relying on textual prompts, ensuring efficient and scalable deployment.

All experiments are conducted using PyTorch \cite{NEURIPS2019_bdbca288} on NVIDIA A100 GPUs. 
We use the Adam optimizer \cite{kingma2014adam}, with the initial learning rate randomly sampled from the range $[10^{-4}, 10^{-2}]$.
%Following the Channel Independence setting in \cite{nie2023time}, each time series channel is modeled independently. 
Following the Channel Independence setting in \cite{nie2023time}, each time series channel is modeled independently. This design simplifies optimization and improves scalability, especially for high-dimensional multivariate forecasting, by allowing TALON to focus on heterogeneous temporal pattern modeling within each variable. We note, however, that such a design may underexploit explicit inter-variable dependencies in applications with strong cross-channel coupling, which remains an important direction for future extension.
The batch size is selected from $\{256, 384\}$, and each model is trained for 10 epochs.
For evaluation, we rerun the baseline models using their official implementations. Specifically, most baselines are obtained from the TimesNet benchmark \cite{wu2023timesnet} and the Timer\_XL repository \cite{liu2025timerxl}. 
For methods not included in these repositories, we follow the original official implementations released by the authors to ensure fair and consistent comparison.

\subsection{Metrics}
\label{sec:metrics}
\textbf{Mean Squared Error (MSE).}
%Mean Squared Error is one of the most widely used metrics for evaluating time series forecasting performance. It calculates the average of the squared differences between predicted values and ground truth values:
Mean Squared Error calculates the average of the squared differences between predicted values and ground truth values:
\begin{align}
    \text{MSE} = \frac{1}{N} \sum_{i=1}^N(y_i - \hat{y}_i)^2.
\end{align}
where $y_i$  and $\hat{y}_i$ denote the true and predicted values, respectively, and 
$N$ is the total number of predictions. %MSE penalizes larger errors more severely, making it sensitive to outliers and suitable for applications that prioritize accurate modeling of extreme values.

\textbf{Mean Absolute Error (MAE).}
Mean Absolute Error measures the average magnitude of the errors between predicted and true values, without considering their direction:
\begin{align}
    \text{MAE} = \frac{1}{N} \sum_{i=1}^N|y_i - \hat{y}_i|.
\end{align}

%Compared to MSE, MAE is more robust to outliers and provides a direct interpretation of the average forecast error in the same units as the original data. It is especially useful when consistent accuracy across the entire forecast range is desired.

%Both MSE and MAE are used in our evaluation to provide a comprehensive assessment of forecasting performance, balancing sensitivity to large deviations (MSE) and overall robustness (MAE).

\textbf{Improvement.}
Improvement (IMP) quantifies the relative performance gain of our proposed method (TALON) over each baseline method. Specifically, it denotes the average percentage reduction in both MSE and MAE across all seven datasets, defined as:
\begin{align}
\text{IMP}_{\text{MSE}} &= \frac{1}{D} \sum_{d=1}^D \frac{\text{MSE}_{\text{baseline}}^{(d)} - \text{MSE}_{\text{TALON}}^{(d)}}{\text{MSE}_{\text{baseline}}^{(d)}}, \\
\text{IMP}_{\text{MAE}} &= \frac{1}{D} \sum_{d=1}^D \frac{\text{MAE}_{\text{baseline}}^{(d)} - \text{MAE}_{\text{TALON}}^{(d)}}{\text{MAE}_{\text{baseline}}^{(d)}},
\end{align}
where $D$ is the number of datasets, and $\text{MSE}_{\text{baseline}}^{(d)}$ and $\text{MAE}_\text{baseline}^{(d)}$ refer to the error metrics of a given baseline on dataset $d$. Positive IMP values indicate that TALON achieves lower errors and thus better forecasting performance.

%IMP provides a concise summary of overall improvement, enabling direct comparison of the relative effectiveness of TALON against each baseline across diverse datasets.

\subsection{Time Series Characteristics}
\label{sec_app:calculation}
We quantify the complexity of each univariate time series segment using three interpretable indicators: trend strength, local variation, and temporal dependency.
Formally, for a univariate segment $s \in \mathbb{R}^{S}$, we extract the following:

\textbf{Trend Strength.}
The trend of a time series refers to the long-term changes or patterns that occur over time. Intuitively, it represents the general direction in which the data is moving. 
Trend strength measures how much of the deseasonalized signal’s variance can be explained by the underlying trend component. 
To compute it, we apply Seasonal-Trend decomposition using Loess (STL) to extract trend, seasonal, and residual components:
\begin{align}
s &= \text{Trend} + \text{Seasonal} + \text{Residual}.
\end{align}
We then calculate the deseasonalized signal $s' = s - \text{Seasonal}$ and define trend strength as:
\begin{align}
\text{TrendStrength} = \max\left(0, 1 - \frac{\mathrm{Var}(\text{Residual})}{\mathrm{Var}(s')} \right).
\end{align}
This formulation reflects the proportion of variance in the deseasonalized signal that is attributable to the trend component.

\textbf{Local Variation.}
We compute the first-order difference $\Delta s_t = s_{t} - s_{t-1}$ and define local variation as:
\begin{align}
    \text{Variation} = \sigma (\text{lag} (1 + \text{std} (\Delta s)) - 1.0),
\end{align}
where $\sigma$ is the sigmoid function. This maps the log-scaled standard deviation to $[0, 1]$ for robust normalization.

\textbf{Temporal Dependency.}
We compute lag-1 autocorrelation:
\begin{align}
    \text{Autocorr} = |\text{acf}(s) [1]|,
\end{align}
where $\text{acf}$ is the autocorrelation function. If the signal is constant or contains invalid values, the score is set to zero for robustness.

The final complexity descriptor is a 3-dimensional vector given by:
\begin{align}
c &= [c_1, c_2, c_3] \\
  &= [\text{TrendStrength}, \text{Variation}, \text{Autocorr}] \in [0,1]^3.
\end{align}
The full procedure for computing the statistical complexity descriptor is outlined in Algorithm~\ref{alg:complexity}.

\begin{algorithm}[ht]
\caption{Statistical Complexity Computation for Time Series Patches}
\label{alg:complexity}
\begin{algorithmic}[1]
\REQUIRE A univariate time series patch $\mathbf{s} \in \mathbb{R}^{S}$
\ENSURE A complexity vector $\mathbf{c} = [c_1, c_2, c_3] \in \mathbb{R}^{3}$

\STATE \textbf{Trend Strength ($c_1$):}
\STATE Apply STL decomposition on $\mathbf{s}$: $\mathbf{s} = \text{Trend} + \text{Seasonal} + \text{Residual}$
\STATE Compute deseasonalized signal: $\mathbf{s}' = \mathbf{s} - \text{Seasonal}$
\IF{$\mathrm{Var}(\mathbf{s}') = 0$}
    \STATE $c_1 \leftarrow 0$
\ELSE
    \STATE $c_1 \leftarrow 1 - \mathrm{Var}(\text{Residual}) / \mathrm{Var}(\mathbf{s}')$
\ENDIF

\STATE \textbf{Derivative Standard Deviation ($c_2$):}
\STATE Compute first-order difference: $\Delta \mathbf{s} = \mathbf{s}_{2:S} - \mathbf{s}_{1:S-1}$
\STATE $c_2 \leftarrow \log(1 + \mathrm{std}(\Delta \mathbf{s}))$, then apply sigmoid scaling: $c_2 \leftarrow 1 / (1 + \exp(-(c_2 - 1.0)))$

\STATE \textbf{Autocorrelation ($c_3$):}
\STATE Compute lag-1 autocorrelation: 
\STATE $c_3 \leftarrow |\mathrm{Corr}(\mathbf{s}_{1:S-1}, \mathbf{s}_{2:S})|$

\RETURN $\mathbf{c} = [c_1, c_2, c_3]$
\end{algorithmic}
\end{algorithm}

\begin{figure*}[t]
    \centering
    \includegraphics[width=0.9\linewidth]{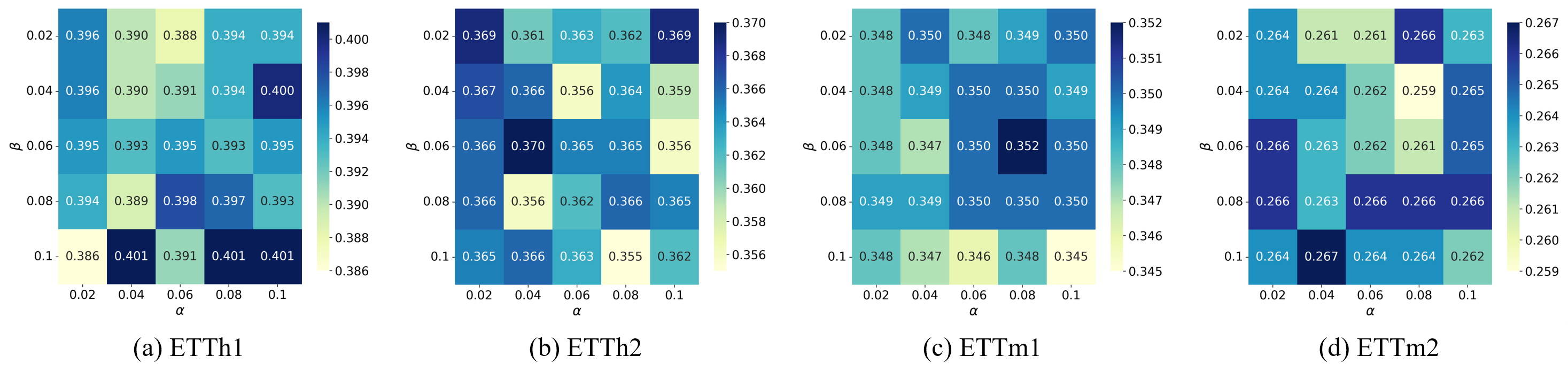}
    \caption{\rev{Parameter sensitivity of $\alpha$ and $\beta$ of the proposed method on the ETTh1, ETTh2, ETTm1, and ETTm2 datasets.}}
    \label{app_fig:Parameter}
\end{figure*}

\begin{table}[t]
\caption{\rev{Parameter sensitivity of $k$ of TALON on the ETTh1, ETTh2, ETTm1, and ETTm2 datasets.}}
\label{tab:topk}
\begin{center}
\resizebox{0.8\linewidth}{!}{
\large
\begin{tabular}{cc|cc|cc|cc}
\toprule
\multicolumn{1}{l|}{}                       & \multicolumn{1}{l|}{} & \multicolumn{2}{c|}{$k=1$}        & \multicolumn{2}{c|}{$k=2$}        & \multicolumn{2}{c}{$k=3$}         \\ \midrule
\multicolumn{1}{c|}{}                       & $H$                     & MSE            & MAE            & MSE            & MAE            & MSE            & MAE            \\ \midrule
\multicolumn{1}{c|}{\multirow{4}{*}{ETTh1}} & 96                    & 0.358          & 0.397          & 0.360          & 0.397          & \textbf{0.351} & \textbf{0.392} \\
\multicolumn{1}{c|}{}                       & 192                   & 0.389          & 0.419          & 0.391          & 0.418          & \textbf{0.381} & \textbf{0.415} \\
\multicolumn{1}{c|}{}                       & 336                   & 0.409          & 0.433          & 0.409          & 0.431          & \textbf{0.398} & \textbf{0.427} \\
\multicolumn{1}{c|}{}                       & 720                   & 0.433          & 0.455          & 0.429          & 0.449          & \textbf{0.414} & \textbf{0.446} \\ \midrule
\rowcolor{gray!30}   
\multicolumn{2}{c|}{Avg.}                                            & 0.397          & 0.426          & 0.397          & 0.424          & \textbf{0.386} & \textbf{0.420} \\ \midrule
\multicolumn{1}{c|}{\multirow{4}{*}{ETTh2}} & 96                    & \textbf{0.282} & \textbf{0.346} & 0.302          & 0.349          & 0.293          & 0.352          \\
\multicolumn{1}{c|}{}                       & 192                   & \textbf{0.345} & 0.390          & 0.355          & \textbf{0.388} & 0.353          & 0.397          \\
\multicolumn{1}{c|}{}                       & 336                   & 0.374          & 0.418          & \textbf{0.371} & \textbf{0.406} & 0.379          & 0.423          \\
\multicolumn{1}{c|}{}                       & 720                   & 0.432          & 0.462          & \textbf{0.393} & \textbf{0.435} & 0.434          & 0.465          \\ \midrule
\rowcolor{gray!30}   
\multicolumn{2}{c|}{Avg.}                                            & 0.358          & 0.404          & \textbf{0.355} & \textbf{0.395} & 0.365          & 0.409          \\ \midrule
\multicolumn{1}{c|}{\multirow{4}{*}{ETTm1}} & 96                    & 0.282          & 0.342          & \textbf{0.278} & \textbf{0.339} & 0.278          & 0.340          \\
\multicolumn{1}{c|}{}                       & 192                   & 0.325          & 0.369          & \textbf{0.324} & \textbf{0.367} & 0.328          & 0.370          \\
\multicolumn{1}{c|}{}                       & 336                   & 0.365          & 0.392          & \textbf{0.358} & \textbf{0.388} & 0.364          & 0.391          \\
\multicolumn{1}{c|}{}                       & 720                   & \textbf{0.413} & \textbf{0.421} & 0.418          & 0.424          & 0.424          & 0.425          \\ \midrule
\rowcolor{gray!30}   
\multicolumn{2}{c|}{Avg.}                                            & 0.346          & 0.381          & \textbf{0.345} & \textbf{0.380} & 0.348          & 0.381          \\ \midrule
\multicolumn{1}{c|}{\multirow{4}{*}{ETTm2}} & 96                    & 0.194          & 0.282          & \textbf{0.172} & \textbf{0.259} & 0.173          & 0.260          \\
\multicolumn{1}{c|}{}                       & 192                   & 0.262          & 0.326          & 0.230          & 0.299          & \textbf{0.223} & \textbf{0.299} \\
\multicolumn{1}{c|}{}                       & 336                   & 0.333          & 0.369          & 0.283          & 0.334          & \textbf{0.278} & \textbf{0.333} \\
\multicolumn{1}{c|}{}                       & 720                   & 0.438          & 0.429          & \textbf{0.361} & 0.385          & 0.362          & \textbf{0.383} \\ \midrule
\rowcolor{gray!30}   
\multicolumn{2}{c|}{Avg.}                                            & 0.307          & 0.352          & 0.261          & 0.319          & \textbf{0.259} & \textbf{0.319} \\ \midrule
\rowcolor{gray!30}   
\multicolumn{2}{c|}{$1^{st}$ Count}                                      & 3              & 2              & \textbf{9}              & \textbf{9}              & 8     & \textbf{9}     \\ \bottomrule
\end{tabular}}
\end{center}
\end{table}

\subsection{Model Analysis}
\label{app_sec:Parameter}

\rev{\textbf{Parameter Sensitivity.}
We assess the effect of $\alpha$ and $\beta$ via a grid search, reporting MSE results on ETTh1, ETTh2, ETTm1, and ETTm2 in Figure~\ref{app_fig:Parameter}.
Across the paired ETT datasets, higher-resolution series (ETTm) generally favor moderate-to-large $\alpha$ values, reflecting the benefit of stronger expert specialization under richer short-term fluctuations, while the hourly ETTh datasets exhibit more diverse preferences (ETTh1 favors smaller $\alpha$, whereas ETTh2 prefers larger $\alpha$), indicating that optimal specialization depends on both sampling resolution and underlying dynamics. 
Moreover, $\beta{=}0.1$ serves as a strong default in most cases, whereas ETTm2 benefits from a smaller $\beta$, suggesting that overly strong alignment can be suboptimal for certain dynamics.}

%As shown in Figure~\ref{app_fig:Parameter}, we further investigate the sensitivity of the hyperparameters $\alpha$ and $\beta$ on three additional datasets: ETTh2, ETTm1, and ETTm2. Across all datasets, our method exhibits strong robustness to a wide range of $\alpha$ and $\beta$ values. The MSE variation across the grid is minimal (mostly within 0.01), indicating stable performance regardless of exact hyperparameter choices. Although slight differences exist in the optimal setting per dataset (e.g., $(\alpha=0.06, \beta=0.06)$ on ETTm1), the overall insensitivity highlights that our method does not depend on meticulous tuning, making it practical and easy to deploy in real-world scenarios.

\rev{In addition, we conduct a sensitivity analysis on the effect of top-$k$ expert selection.
As shown in Table~\ref{tab:topk}, both $k=2$ and $k=3$ achieve competitive performance across most datasets and prediction lengths.
Specifically, $k=2$ yields the best results overall, while $k=3$ also performs strongly.
This suggests that leveraging multiple experts generally improves the model's ability to capture heterogeneous temporal patterns, compared to using a single expert ($k=1$).
Moreover, the performance remains relatively stable across different $k$ values, demonstrating the robustness of the routing mechanism.}

\begin{table}[t]
\caption{Effect of prompt field ordering.}
\label{tab:ordering}
\begin{center}
\resizebox{1.0\linewidth}{!}{
\large
\begin{tabular}{cc|cc|cc|cc|cc|cc|cc}
\toprule
\multicolumn{2}{c|}{Models}                        & \multicolumn{2}{c|}{TALON}      & \multicolumn{2}{c|}{Prompt 2}   & \multicolumn{2}{c|}{Prompt 3}   & \multicolumn{2}{c|}{Prompt 4}   & \multicolumn{2}{c|}{Prompt 5} & \multicolumn{2}{c}{Prompt 6} \\ \midrule
\multicolumn{2}{c|}{Metric}                        & MSE            & MAE            & MSE            & MAE            & MSE            & MAE            & MSE            & MAE            & MSE           & MAE           & MSE           & MAE          \\ \midrule
\multicolumn{1}{c|}{}                        & 96  & \textbf{0.351} & \textbf{0.392} & \underline{0.352}    & 0.393          & 0.354          & 0.394          & 0.353          & {\underline{0.392}}    & 0.355         & 0.393         & 0.355         & 0.395        \\
\multicolumn{1}{c|}{}                        & 192 & \textbf{0.381} & \underline{0.415}    & 0.384          & 0.415          & 0.383          & 0.415          & \underline{0.383}    & \textbf{0.414} & 0.386         & 0.415         & 0.387         & 0.417        \\
\multicolumn{1}{c|}{}                        & 336 & \textbf{0.398} & \textbf{0.427} & 0.402          & 0.428          & \underline{ 0.399}    & \underline{0.427}    & 0.401          & 0.428          & 0.404         & 0.429         & 0.404         & 0.430        \\
\multicolumn{1}{c|}{\multirow{-4}{*}{ETTh1}} & 720 & \textbf{0.414} & \underline{0.446}    & 0.425          & 0.450          & \underline{0.416}    & 0.446          & 0.418          & \textbf{0.445} & 0.426         & 0.451         & 0.427         & 0.451        \\ \midrule
\rowcolor{gray!30}
\multicolumn{2}{c|}{\cellcolor[HTML]{D0CECE}Avg.}  & \textbf{0.386} & \textbf{0.420} & 0.391          & 0.421          & \underline{0.388}    & \underline{0.420}    & 0.389          & 0.420          & 0.393         & 0.422         & 0.393         & 0.423        \\ \midrule
\multicolumn{1}{c|}{}                        & 96  & 0.302          & 0.349          & \underline{0.284}    & \textbf{0.342} & 0.292          & 0.348          & \textbf{0.280} & \underline{0.344}    & 0.290         & 0.346         & 0.288         & 0.348        \\
\multicolumn{1}{c|}{}                        & 192 & 0.355          & \underline{0.388}    & \underline{0.349}    & \textbf{0.385} & 0.355          & 0.391          & \textbf{0.343} & 0.388          & 0.352         & 0.389         & 0.350         & 0.391        \\
\multicolumn{1}{c|}{}                        & 336 & \underline{0.371}    & \underline{0.406}    & \textbf{0.368} & \textbf{0.405} & 0.376          & 0.412          & 0.371          & 0.412          & 0.373         & 0.412         & 0.372         & 0.412        \\
\multicolumn{1}{c|}{\multirow{-4}{*}{ETTh2}} & 720 & \textbf{0.393} & \textbf{0.435} & \underline{0.404}    & \underline{0.439}    & 0.416          & 0.447          & 0.421          & 0.452          & 0.422         & 0.452         & 0.415         & 0.449        \\ \midrule
\rowcolor{gray!30}
\multicolumn{2}{c|}{\cellcolor[HTML]{D0CECE}Avg.}  & 0.355          & \underline{0.395}    & \textbf{0.352} & \textbf{0.393} & 0.360          & 0.399          & \underline{0.354}    & 0.399          & 0.359         & 0.400         & 0.356         & 0.400        \\ \midrule
\multicolumn{1}{c|}{}                        & 96  & \textbf{0.278} & 0.339          & 0.281          & 0.340          & 0.279          & \textbf{0.338} & \underline{0.279}    & \underline{0.338}    & 0.282         & 0.340         & 0.287         & 0.343        \\
\multicolumn{1}{c|}{}                        & 192 & \textbf{0.324} & \textbf{0.367} & 0.326          & 0.369          & \underline{0.325}    & 0.367          & 0.326          & \underline{0.367}    & 0.329         & 0.369         & 0.334         & 0.371        \\
\multicolumn{1}{c|}{}                        & 336 & \textbf{0.358} & \textbf{0.388} & 0.360          & 0.390          & \underline{0.359}    & 0.388          & 0.361          & \underline{0.388}    & 0.364         & 0.391         & 0.371         & 0.393        \\
\multicolumn{1}{c|}{\multirow{-4}{*}{ETTm1}} & 720 & 0.418          & 0.424          & \textbf{0.414} & 0.422          & \underline{0.415}    & \textbf{0.421} & 0.416          & \underline{0.421}    & 0.420         & 0.424         & 0.430         & 0.426        \\ \midrule
\rowcolor{gray!30}
\multicolumn{2}{c|}{\cellcolor[HTML]{D0CECE}Avg.}  & \textbf{0.345} & 0.380          & 0.345          & 0.380          & \underline{0.345}    & \textbf{0.378} & 0.345          & \underline{0.379}    & 0.349         & 0.381         & 0.355         & 0.383        \\ \midrule
\multicolumn{1}{c|}{}                        & 96  & \underline{0.173}    & \textbf{0.260} & 0.174          & 0.261          & \textbf{0.172} & \underline{0.260}    & 0.174          & 0.261          & 0.176         & 0.263         & 0.175         & 0.260        \\
\multicolumn{1}{c|}{}                        & 192 & \textbf{0.223} & \textbf{0.299} & \underline{0.231}    & \underline{0.300}    & 0.231          & 0.300          & 0.232          & 0.300          & 0.234         & 0.302         & 0.234         & 0.300        \\
\multicolumn{1}{c|}{}                        & 336 & \textbf{0.278} & \textbf{0.333} & \underline{0.282}    & \underline{0.334}    & 0.284          & 0.335          & 0.284          & 0.335          & 0.287         & 0.337         & 0.286         & 0.335        \\
\multicolumn{1}{c|}{\multirow{-4}{*}{ETTm2}} & 720 & \underline{0.362}    & \textbf{0.383} & \textbf{0.360} & \underline{0.384}    & 0.363          & 0.387          & 0.366          & 0.387          & 0.367         & 0.388         & 0.363         & 0.385        \\ \midrule
\rowcolor{gray!30}
\multicolumn{2}{c|}{\cellcolor[HTML]{D0CECE}Avg.}  & \textbf{0.259} & \textbf{0.319} & \underline{0.262}    & \underline{0.320}    & 0.262          & 0.320          & 0.264          & 0.321          & 0.266         & 0.323         & 0.264         & 0.320        \\ \bottomrule
\end{tabular}}
\end{center}
\end{table}

\textbf{Effect of Prompt Field Ordering.}
To evaluate the sensitivity of the proposed representation alignment module to prompt structure, we constructed several alternative token-adaptive templates by rearranging the order of the three semantic components—expert routing hints, patch-wise temporal context, and complexity-related features—while keeping their content unchanged. The variants are defined as follows: 
\begin{itemize}
    \item Prompt 2: patch-wise temporal context, expert routing hints, complexity-related features
    \item Prompt 3: expert routing hints, complexity-related features, patch-wise temporal context
    \item Prompt 4: complexity-related features, expert routing hints, patch-wise temporal context
    \item Prompt 5: patch-wise temporal context, complexity-related features, expert routing hints
    \item Prompt 6: complexity-related features, patch-wise temporal context, expert routing hints
\end{itemize}
As shown in Table~\ref{tab:ordering}, TALON exhibits stable performance across these variants, whereas the original prompt consistently yields the best results. In particular, placing the numerical complexity-related features at the end achieves the strongest performance, consistent with observations in TimeCMA \cite{liu2025timecma}. These results indicate that alignment effectiveness primarily stems from the structured semantic fields, while the ordering of these fields subtly influences final model performance. Overall, this demonstrates that TALON’s representation alignment module is robust to variations in prompt phrasing.

\begin{table}[t]
\caption{Comparison of TALON with augmented routing.}
\label{tab:routing}
\begin{center}
\resizebox{0.8\linewidth}{!}{
\large
\begin{tabular}{cc|cc|cc|cc}
\toprule
\multicolumn{2}{c|}{Models}                        & \multicolumn{2}{c|}{TALON}      & \multicolumn{2}{c|}{TALON + mean} & \multicolumn{2}{c}{TALON + mean/std} \\ \midrule
\multicolumn{2}{c|}{Metric}                        & MSE            & MAE            & MSE            & MAE           & MSE               & MAE              \\ \midrule
\multicolumn{1}{c|}{}                        & 96  & \textbf{0.351} & \textbf{0.392} & 0.361          & 0.396         & 0.359             & 0.393            \\
\multicolumn{1}{c|}{}                        & 192 & \textbf{0.381} & \textbf{0.415} & 0.393          & 0.418         & 0.390             & 0.416            \\
\multicolumn{1}{c|}{}                        & 336 & \textbf{0.398} & \textbf{0.427} & 0.412          & 0.431         & 0.408             & 0.429            \\
\multicolumn{1}{c|}{\multirow{-4}{*}{ETTh1}} & 720 & \textbf{0.414} & \textbf{0.446} & 0.445          & 0.458         & 0.434             & 0.453            \\ \midrule
\rowcolor{gray!30}  
\multicolumn{2}{c|}{Avg.}  & \textbf{0.386} & \textbf{0.420} & 0.403          & 0.426         & 0.398             & 0.423            \\ \midrule
\multicolumn{1}{c|}{}                        & 96  & \textbf{0.302} & \textbf{0.349} & 0.305          & 0.353         & 0.302             & 0.355            \\
\multicolumn{1}{c|}{}                        & 192 & \textbf{0.355} & \textbf{0.388} & 0.356          & 0.391         & 0.358             & 0.398            \\
\multicolumn{1}{c|}{}                        & 336 & \textbf{0.371} & \textbf{0.406} & 0.373          & 0.410         & 0.380             & 0.421            \\
\multicolumn{1}{c|}{\multirow{-4}{*}{ETTh2}} & 720 & \textbf{0.393} & \textbf{0.435} & 0.399          & 0.439         & 0.447             & 0.470            \\ \midrule
\rowcolor{gray!30}  
\multicolumn{2}{c|}{Avg.}  & \textbf{0.355} & \textbf{0.395} & 0.358          & 0.399         & 0.372             & 0.411            \\ \midrule
\multicolumn{1}{c|}{}                        & 96  & 0.278          & 0.339          & 0.279          & 0.339         & \textbf{0.277}    & \textbf{0.339}   \\
\multicolumn{1}{c|}{}                        & 192 & 0.324          & 0.367          & 0.327          & 0.368         & \textbf{0.322}    & \textbf{0.367}   \\
\multicolumn{1}{c|}{}                        & 336 & 0.358          & \textbf{0.388} & 0.362          & 0.390         & \textbf{0.357}    & 0.389            \\
\multicolumn{1}{c|}{\multirow{-4}{*}{ETTm1}} & 720 & 0.418          & 0.424          & 0.415          & 0.423         & \textbf{0.413}    & \textbf{0.423}   \\ \midrule
\rowcolor{gray!30}  
\multicolumn{2}{c|}{Avg.}  & 0.345          & 0.380          & 0.346          & 0.380         & \textbf{0.342}    & \textbf{0.379}   \\ \midrule
\multicolumn{1}{c|}{}                        & 96  & \textbf{0.173} & \textbf{0.260} & 0.177          & 0.263         & 0.178             & 0.264            \\
\multicolumn{1}{c|}{}                        & 192 & \textbf{0.223} & \textbf{0.299} & 0.236          & 0.303         & 0.237             & 0.304            \\
\multicolumn{1}{c|}{}                        & 336 & \textbf{0.278} & \textbf{0.333} & 0.290          & 0.338         & 0.292             & 0.339            \\
\multicolumn{1}{c|}{\multirow{-4}{*}{ETTm2}} & 720 & \textbf{0.362} & \textbf{0.383} & 0.374          & 0.390         & 0.375             & 0.391            \\ \midrule
\rowcolor{gray!30}  
\multicolumn{2}{c|}{Avg.}  & \textbf{0.259} & \textbf{0.319} & 0.269          & 0.324         & 0.270             & 0.324            \\ \bottomrule
\end{tabular}}
\end{center}
\end{table}

\begin{figure*}[t]
    \centering
        \includegraphics[width=0.23\linewidth]{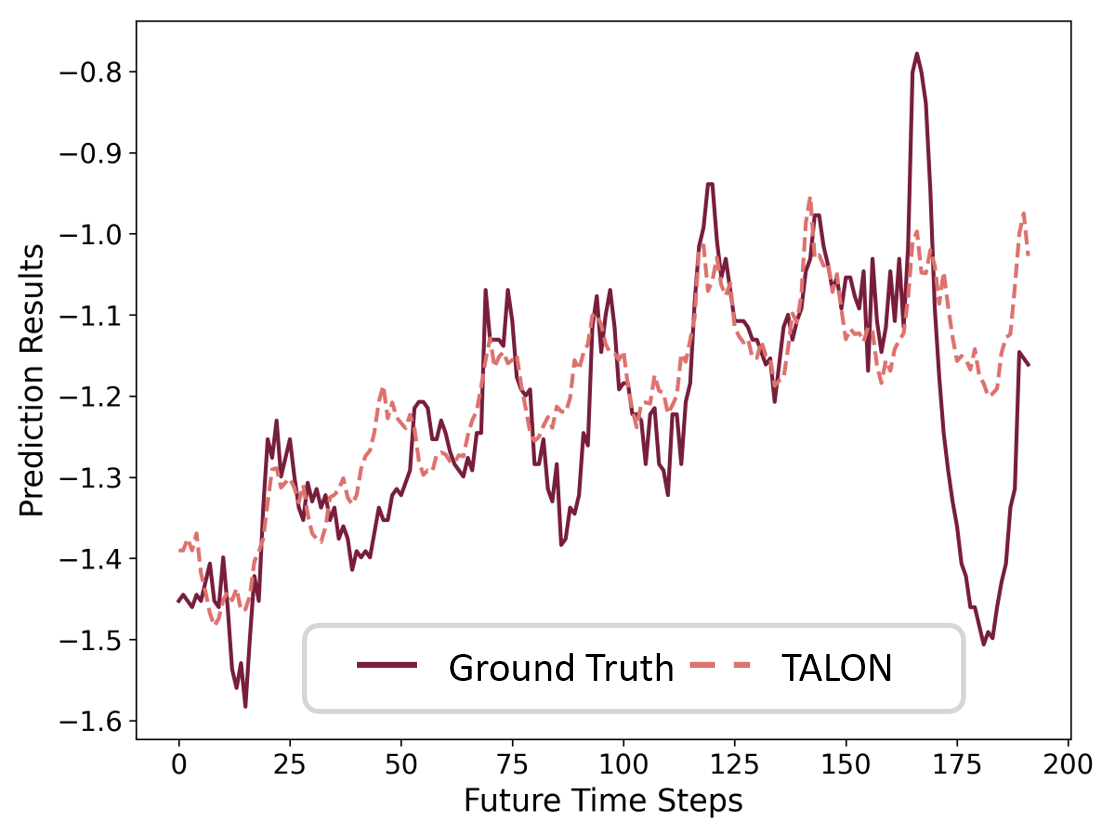}\hspace{0.01\linewidth} 
        \includegraphics[width=0.23\linewidth]{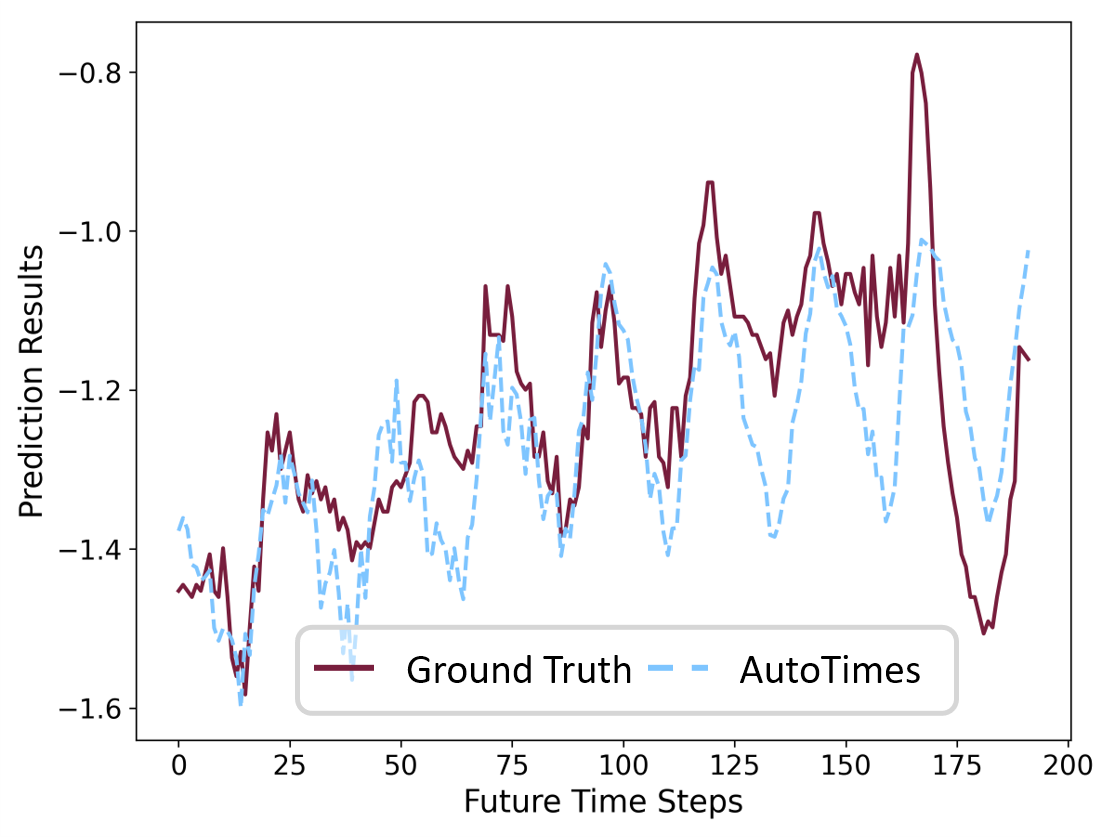}\hspace{0.01\linewidth} 
        \includegraphics[width=0.23\linewidth]{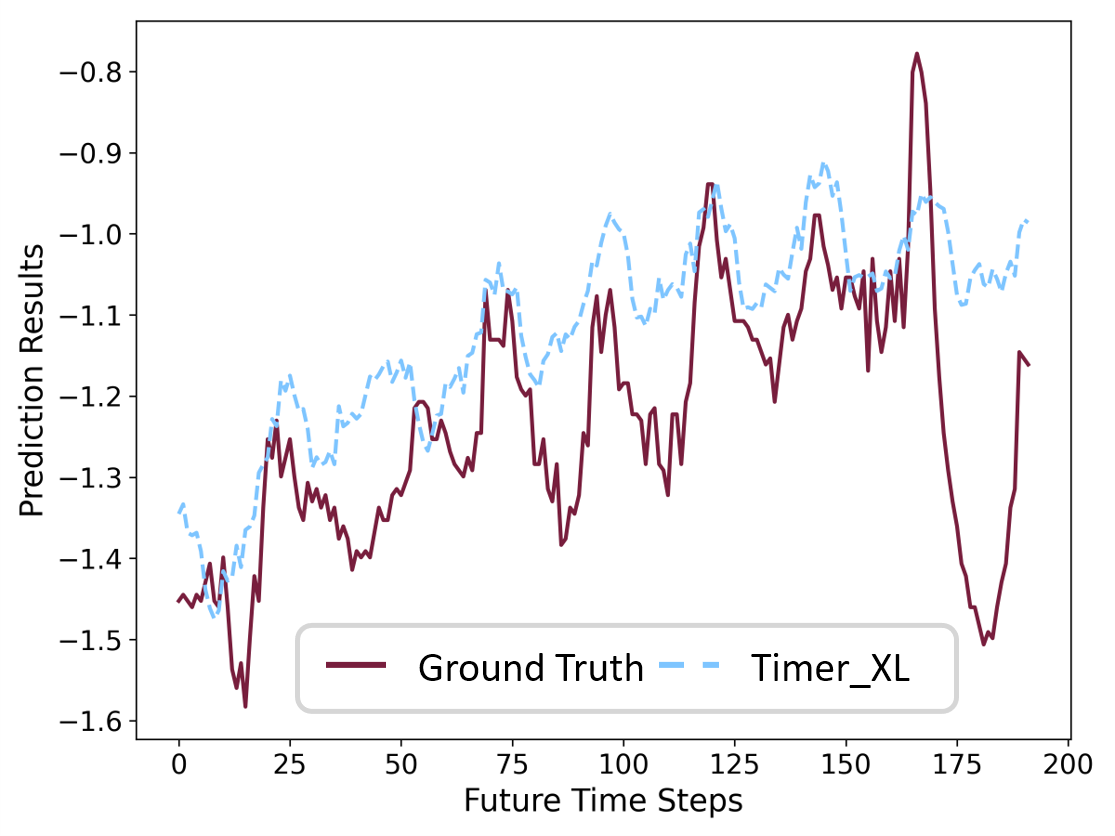}\hspace{0.01\linewidth} 
        \includegraphics[width=0.23\linewidth]{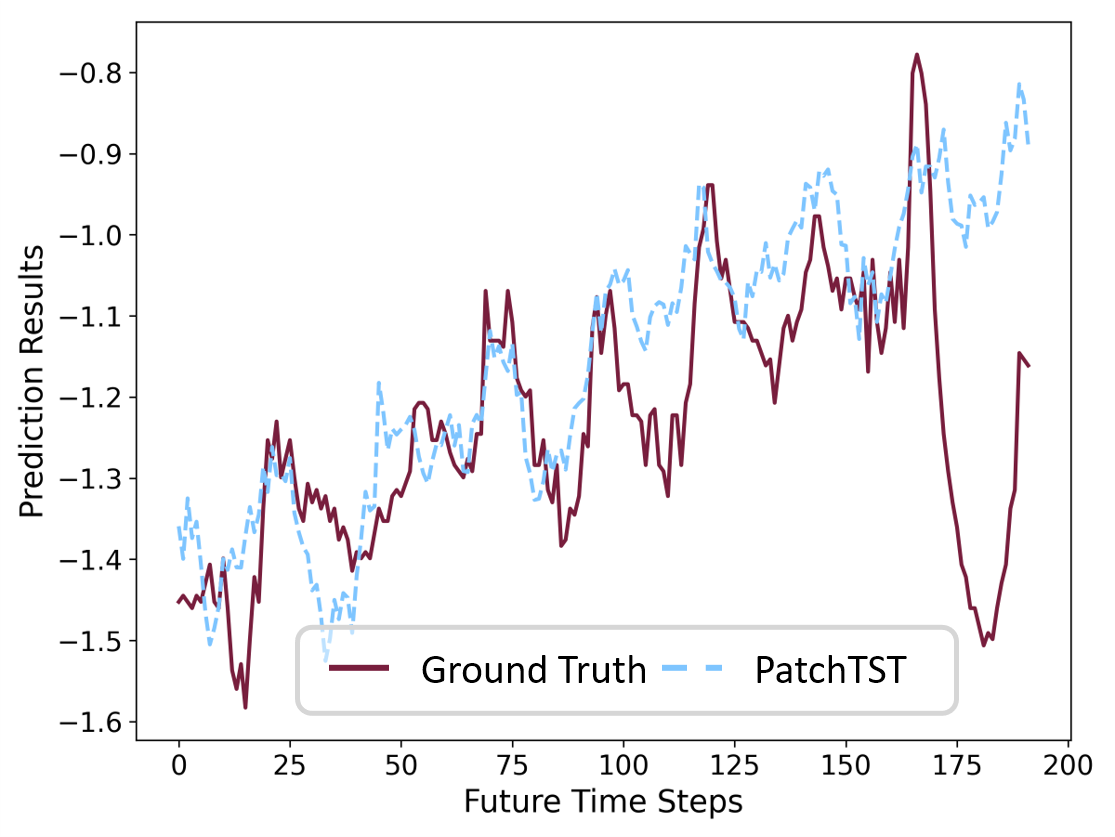}
        \caption{Forecasting examples across ETTh1 (672-step input, 192-step prediction).}
    \label{fig:forecast-examples-ETTh1}
\end{figure*}

\begin{figure*}
    \centering
        \includegraphics[width=0.23\linewidth]{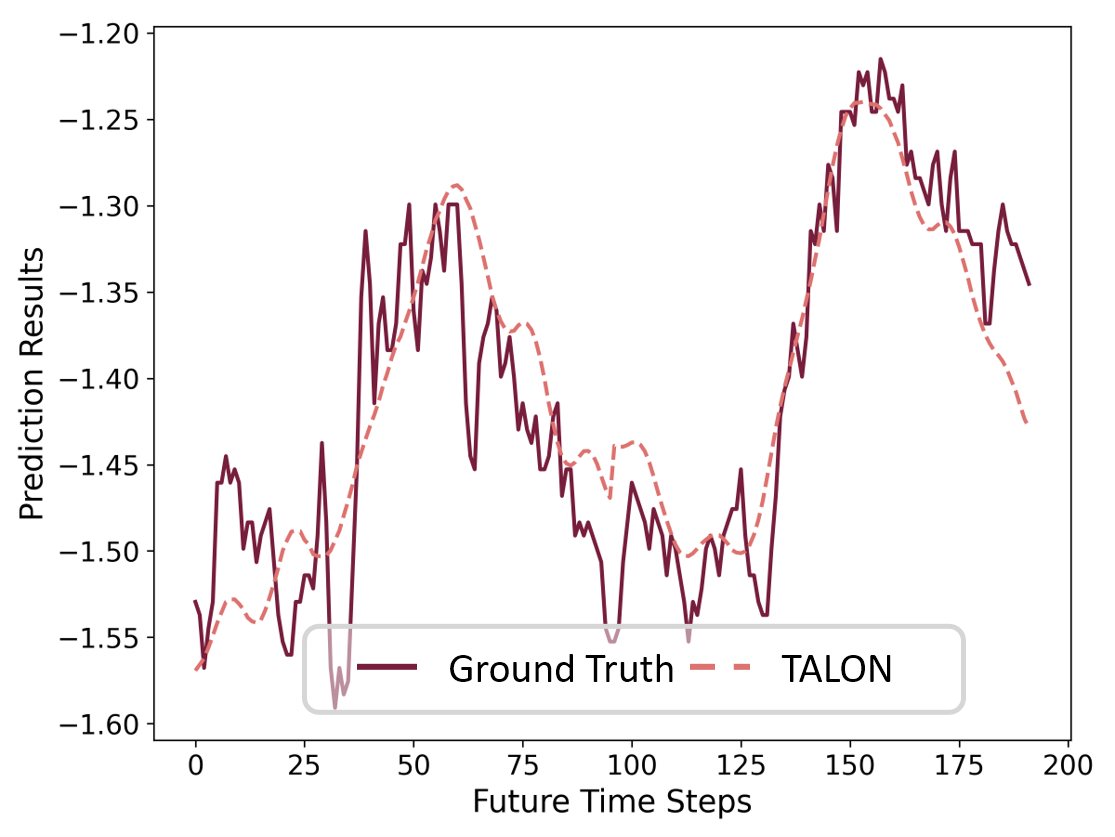}\hspace{0.01\linewidth} 
        \includegraphics[width=0.23\linewidth]{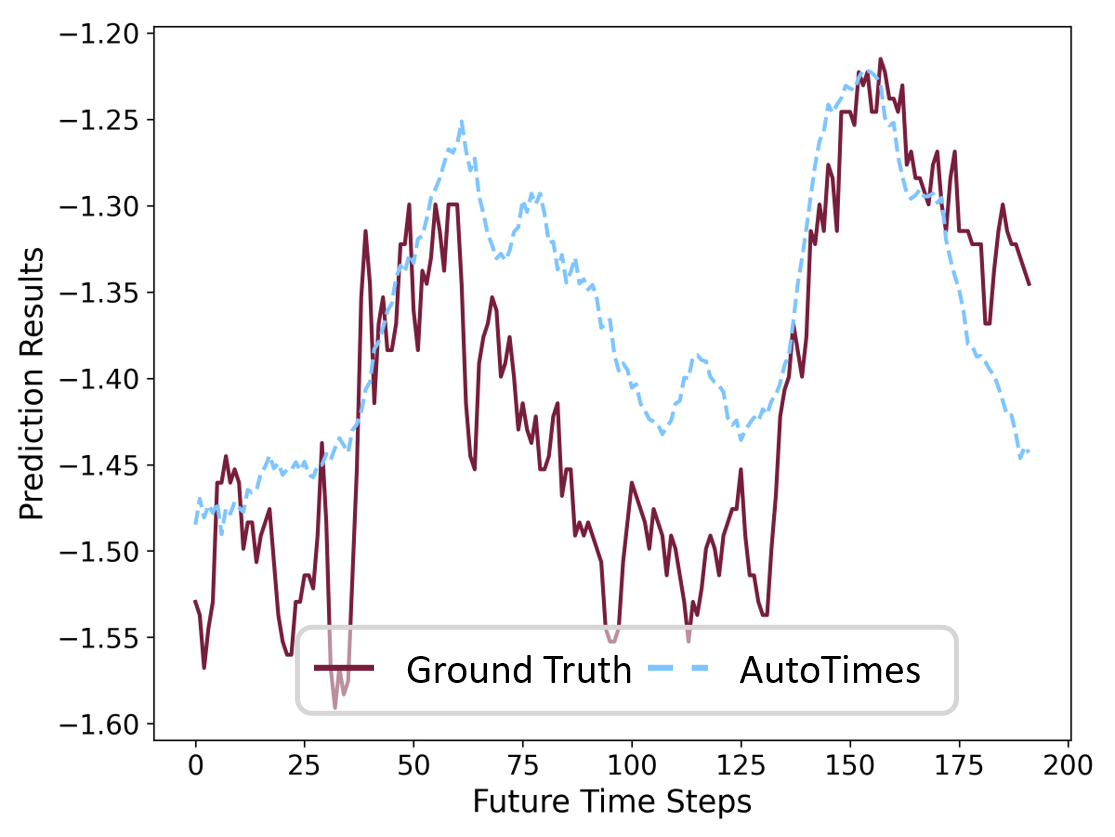}\hspace{0.01\linewidth} 
        \includegraphics[width=0.23\linewidth]{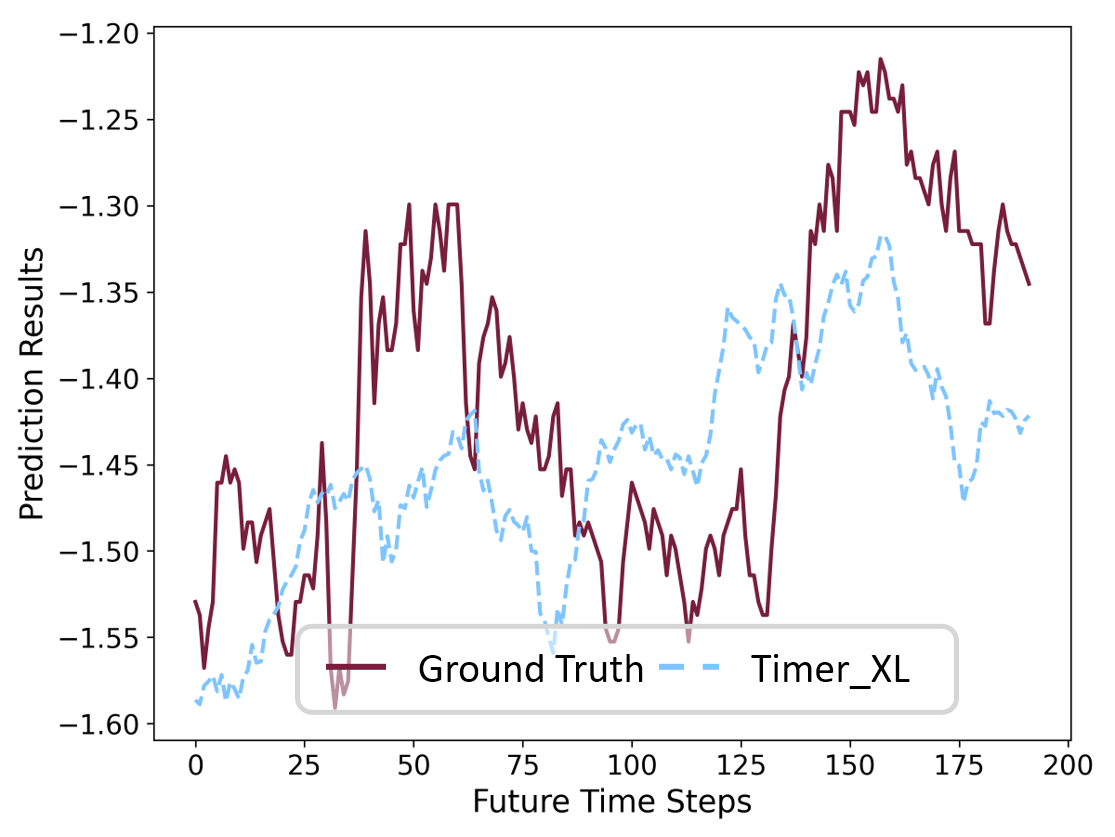}\hspace{0.01\linewidth} 
        \includegraphics[width=0.23\linewidth]{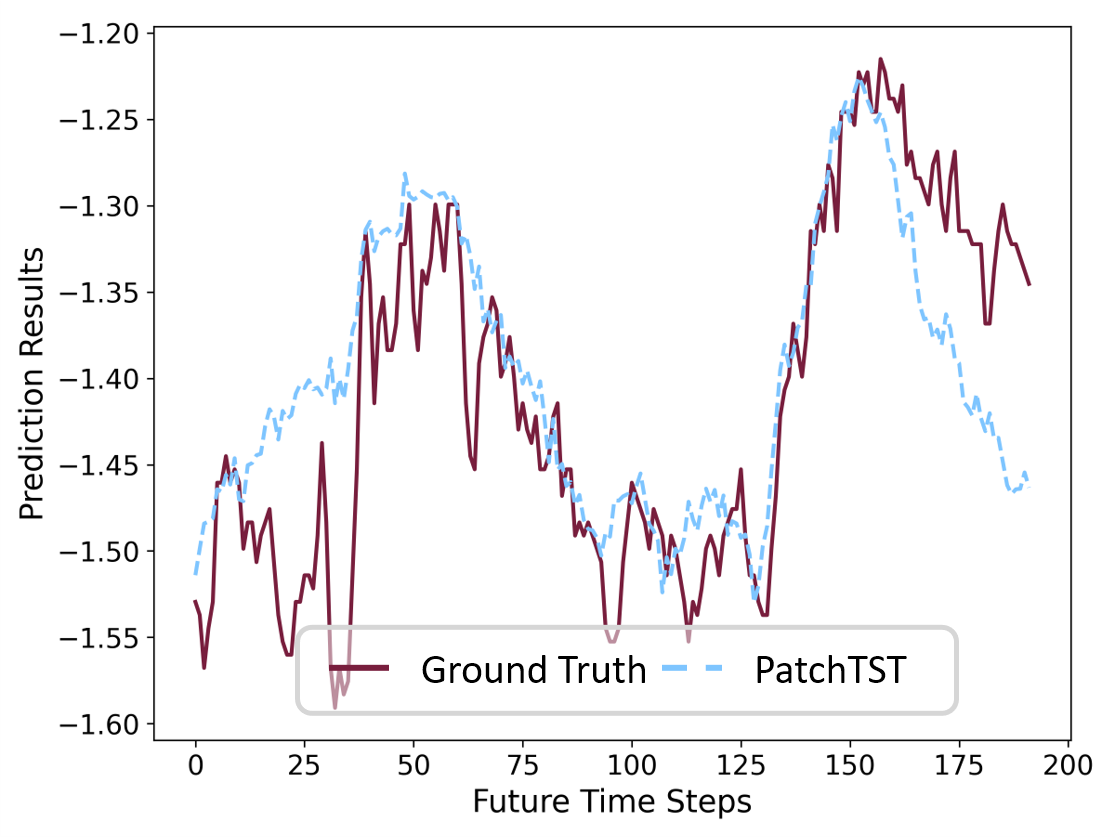}
        \caption{Forecasting examples across ETTm1 (672-step input, 192-step prediction).}
    \label{fig:forecast-examples-ETTm1}
\end{figure*}

\textbf{Comparison with Richer Routing Statistics.}
To further examine whether richer statistical descriptors improve pattern quantification, we augment the original three descriptors in TALON with additional statistics, including mean and standard deviation. As shown in Table~\ref{tab:routing}, these augmented variants do not consistently outperform the original TALON design across datasets and forecasting horizons. While slight improvements can be observed in a few cases, the original TALON design remains more stable overall and delivers the most competitive average performance on most datasets. These findings suggest that the proposed three descriptors already provide sufficiently informative and complementary routing cues for expert selection, supporting our choice of a compact descriptor set with good effectiveness, simplicity, and interpretability. Nevertheless, the current design is intentionally lightweight, and further gains may require richer or learned routing statistics when handling more structurally complex temporal patterns.

%\begin{figure*}
%    \centering
%        \includegraphics[width=0.23\linewidth]{Figure/weather_TALON_192.png}\hspace{0.01\linewidth} 
%        \includegraphics[width=0.23\linewidth]{Figure/weather_AutoTimes_192.png}\hspace{0.01\linewidth} 
%        \includegraphics[width=0.23\linewidth]{Figure/weather_Timer_xl_192.png}\hspace{0.01\linewidth} 
%        \includegraphics[width=0.23\linewidth]{Figure/weather_PatchTST_192.png}
%        \caption{Forecasting examples across Weather (672-step input, 192-step prediction).}
%    \label{fig:forecast-examples-Weather}
%\end{figure*}

\rev{\subsection{Showcases}}
\label{app_sec:visualization}
To further illustrate the forecasting quality of TALON, we randomly select representative prediction examples from two additional datasets: ETTm1 and Weather, each with a forecast horizon of 192 time steps.
We compare TALON against three strong baselines: Timer\_XL \cite{liu2025timerxl}, AutoTimes \cite{liu2024autotimes}, and PatchTST \cite{nie2023time}.
As shown in Figures~\ref{fig:forecast-examples-ETTh1}--\ref{fig:forecast-examples-ETTm1}, TALON consistently produces predictions that better align with the ground truth, particularly in segments exhibiting nonstationarity, local fluctuations, or abrupt structural shifts.

These improvements stem from TALON’s Heterogeneous Temporal Encoder, which employs a mixture of diverse architectural primitives to accommodate varying levels of temporal complexity. This design allows TALON to flexibly capture sharp transitions, smooth trends, and localized irregularities, avoiding the modeling bias introduced by homogeneous structures.
In contrast, methods like PatchTST and AutoTimes often rely on fixed patch tokenization or prompt-based representations, which may be less robust when faced with irregular periodicities or regime shifts.

Furthermore, TALON leverages pretrained LLMs to produce structured reference embeddings from natural-language descriptions of patch-level statistical cues. These prompt embeddings serve as informative priors that enhance the model’s understanding of temporal structure.
By incorporating natural language priors, TALON gains a higher-level understanding of variable dependencies and temporal structures, which is especially beneficial in noisy or nonstationary environments.
%Overall, these qualitative results support TALON’s design philosophy of pattern-aware forecasting with representation-level alignment, demonstrating robust generalization across diverse datasets and dynamic regimes.

\subsection{Broader Impact}
\textbf{Impact on Real-world Applications.}
By aligning statistical time series representations with the representational space of large language models, TALON enables effective utilization of LLM capabilities without relying on textual prompts or auxiliary language inputs during inference. This prompt-free design makes TALON particularly suitable for real-world forecasting scenarios where rich textual context is unavailable, unreliable, or costly to maintain. 
In domains such as energy systems, finance, and supply chain management, time series data often exhibit strong nonstationarity and evolving local patterns, while operational constraints require low-latency and low-maintenance deployment. TALON addresses these challenges by providing a scalable and adaptable forecasting framework that generalizes across dynamic regimes without frequent retraining, thereby reducing operational cost and improving robustness in practical settings.

\textbf{Impact on Future Research.}
Beyond empirical performance, TALON points to a representation-centric paradigm for leveraging large language models in time series forecasting. 
By demonstrating that aligning temporal representations with the latent space of LLMs is sufficient to unlock their sequence modeling capabilities, TALON shifts the focus from explicit language interfaces, such as tokenization or prompt engineering, toward principled representation alignment.
This perspective opens new research directions on understanding how temporal dynamics map onto pretrained language model representations, potentially bridging time series analysis with foundation model theory.
Moreover, TALON’s autoregressive formulation enables horizon-agnostic forecasting, encouraging future work on train-once, infer-many evaluation protocols and scalable forecasting models that generalize across variable prediction lengths.

%全文不超过10页

\bibliographystyle{IEEEtran}
\bibliography{reference}

\end{document}